%% file: neurips_2026.tex
\documentclass{article}

 \usepackage[preprint]{neurips_2026}

\usepackage[utf8]{inputenc} 
\usepackage[T1]{fontenc}    
\usepackage{hyperref}       
\usepackage{url}            
\usepackage{booktabs}       
\usepackage{amsfonts}       
\usepackage{nicefrac}       
\usepackage{microtype}      
\usepackage{xcolor}   

\usepackage{hyperref}
\usepackage{url}

\usepackage{amsfonts}       
\usepackage{nicefrac}       
\usepackage{microtype}      
\usepackage{xcolor}         
\usepackage{amsmath}
\usepackage{amsfonts}
\usepackage{amssymb}
\usepackage{amsthm}
\usepackage{enumitem}
\usepackage[none]{hyphenat}
\usepackage{hyperref}
\usepackage{xcolor}
\usepackage{graphicx}
\usepackage[most]{tcolorbox}
\usepackage{natbib}
\usepackage{subcaption} 
\usepackage{tabularx}
\usepackage{booktabs}

\usepackage{graphicx}
\usepackage{minted} 
\usepackage{multirow}
\usepackage{booktabs}
\usepackage{amsfonts}
\usepackage{wrapfig}
\usepackage{float}
\usepackage[most]{tcolorbox}

\usepackage{pgfplots}
\pgfplotsset{compat=1.18}
\usepackage{xcolor}

\usepackage{pgfplots}

\usepackage[table]{xcolor}

\pgfplotsset{compat=1.18}
\usetikzlibrary{matrix}

\definecolor{newblue}{HTML}{0064E0}

\definecolor{nblue}{HTML}{E3EDFF}

\definecolor{ngreen}{HTML}{F3F1E8}

\definecolor{nred}{HTML}{FFE8ED}

\newcommand{\deltag}[2]{\cellcolor{green!#2}{\textbf{+}#1}}
\newcommand{\deltar}[2]{\cellcolor{red!#2}{\textbf{--}#1}}
\newcommand{\deltaz}{\cellcolor{gray!10}{--}}

\newcommand{\oursrow}{\rowcolor{blue!6}} 

\hypersetup{
  colorlinks,
  linkcolor=newblue,
  citecolor=newblue,
  urlcolor=newblue
}

\renewcommand{\arraystretch}{1.1}

\newtcolorbox[auto counter]{qoutebox}[2][]{%
  colframe=ngreen, 
  colback=ngreen,  
  #1
}

\newtcolorbox[auto counter]{algorithmbox}[2][]{%
  colframe=purple!10!white,     
  colback=purple!10!white,      
  before upper={{\textbf{Algorithm \thetcbcounter.} #2}}, 
  #1
}

\newtcolorbox[auto counter]{mainbox}[2][]{%
  colframe=ngreen, 
  colback=ngreen,  
  breakable,
  before upper={{\textbf{Main Contributions.} #2}},
  #1
}

\newtcolorbox[auto counter]{definitionbox}[2][]{%
  colframe=nblue,, 
  colback=nblue,  
  breakable,
  before upper={{\textbf{Definition \thetcbcounter.} #2}},
  #1
}

\makeatletter
\def\tcb@cnt@definitionboxautorefname{Def.}
\makeatother

\newcommand{\method}{\textsc{Murphy}}

\usepackage[capitalize,noabbrev]{cleveref}
\usepackage[textsize=tiny]{todonotes}

\title{\textsc{Murphy}: Feedback-Aware GRPO with Retrospective Credit Assignment for Multi-Turn Code Generation}

\author{%
  Chanakya Ekbote\thanks{Equal contribution and corresponding author(s). Chanakya Ekbote was an intern at AWS AI during this work.}
  \textsuperscript{   \includegraphics[height=1.5em]{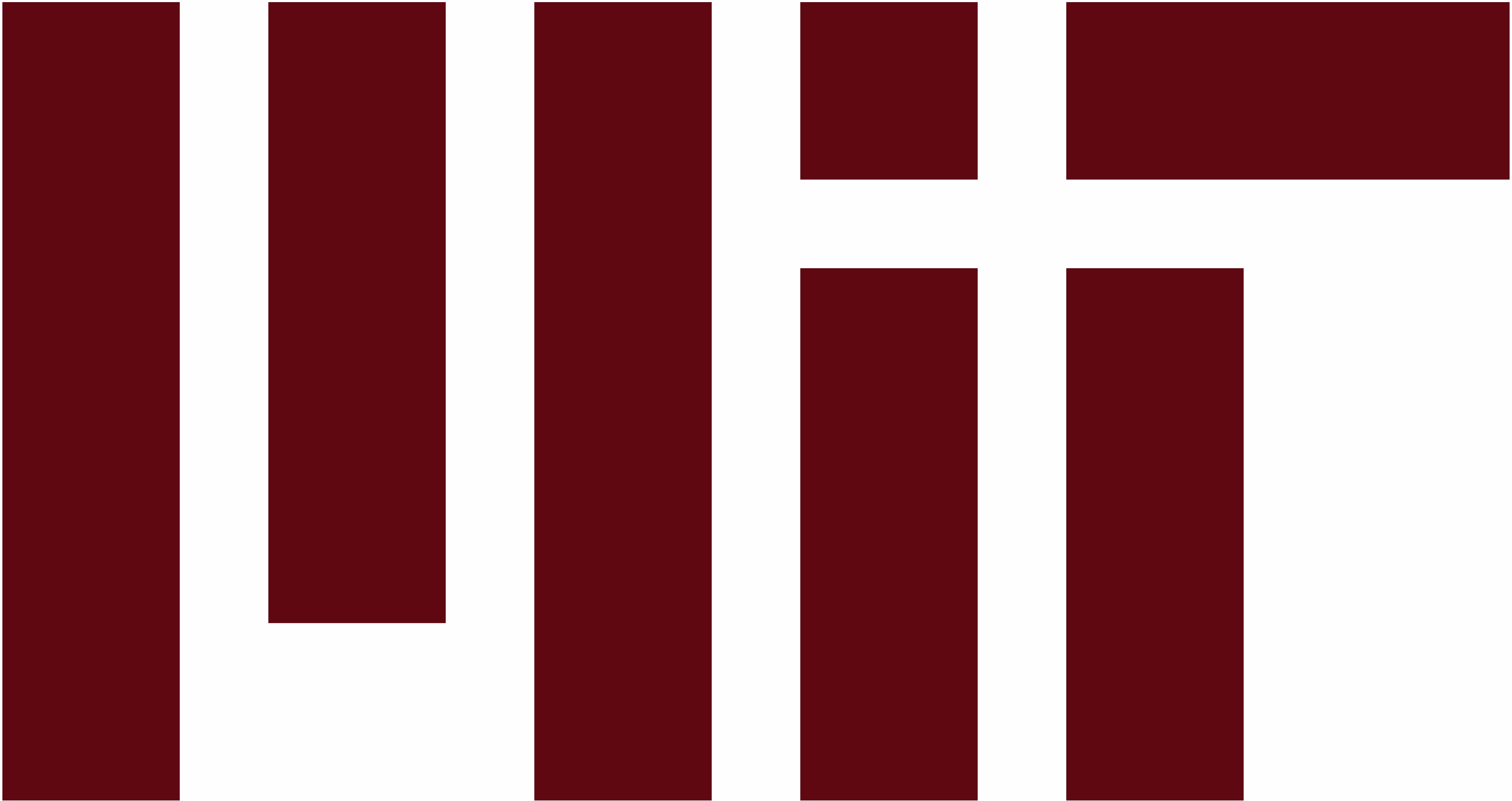}}
  \\
  \href{mailto:cekbote@mit.edu}{\texttt{cekbote@mit.edu}}
  \\
  \And
  Vijay Lingam\footnotemark[\value{footnote}]
  \textsuperscript{   \includegraphics[height=1.8em]{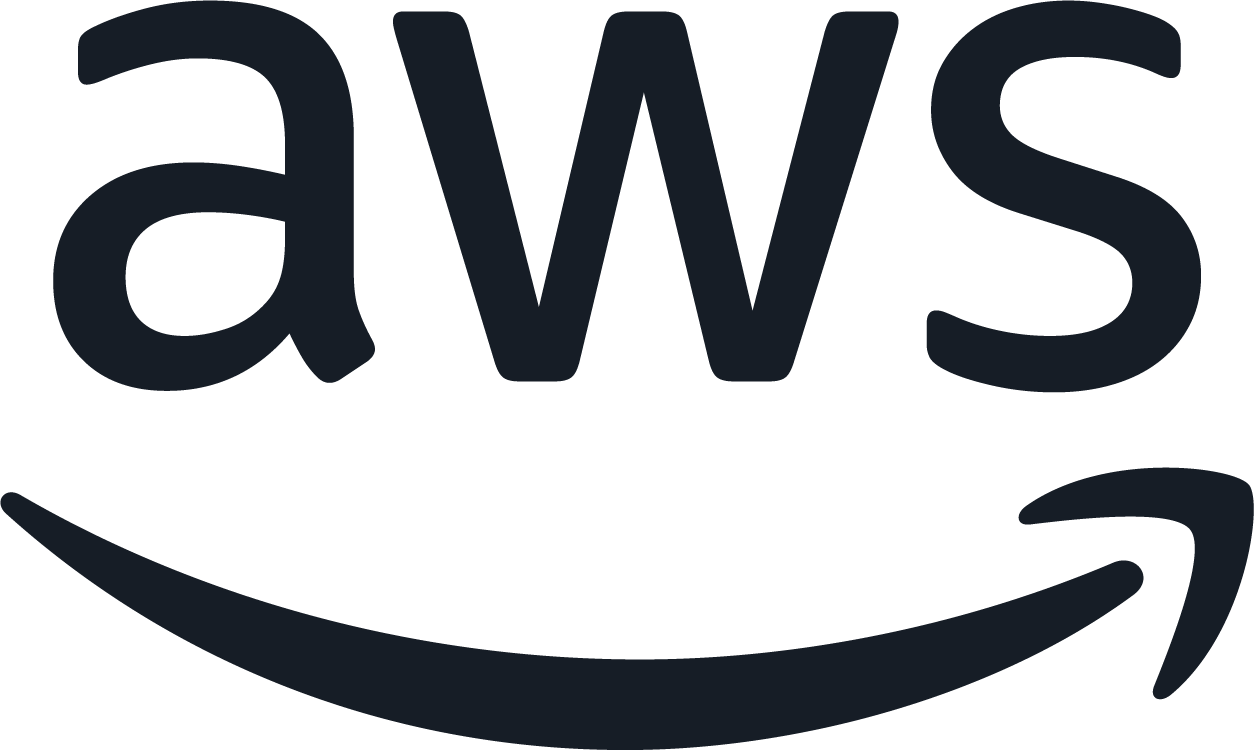}}
  \\
  \href{mailto:vlingam@amazon.com}{\texttt{vlingam@amazon.com}}
  \\
  \AND
  Sujay Sanghavi
  \textsuperscript{ \includegraphics[height=1.5em]{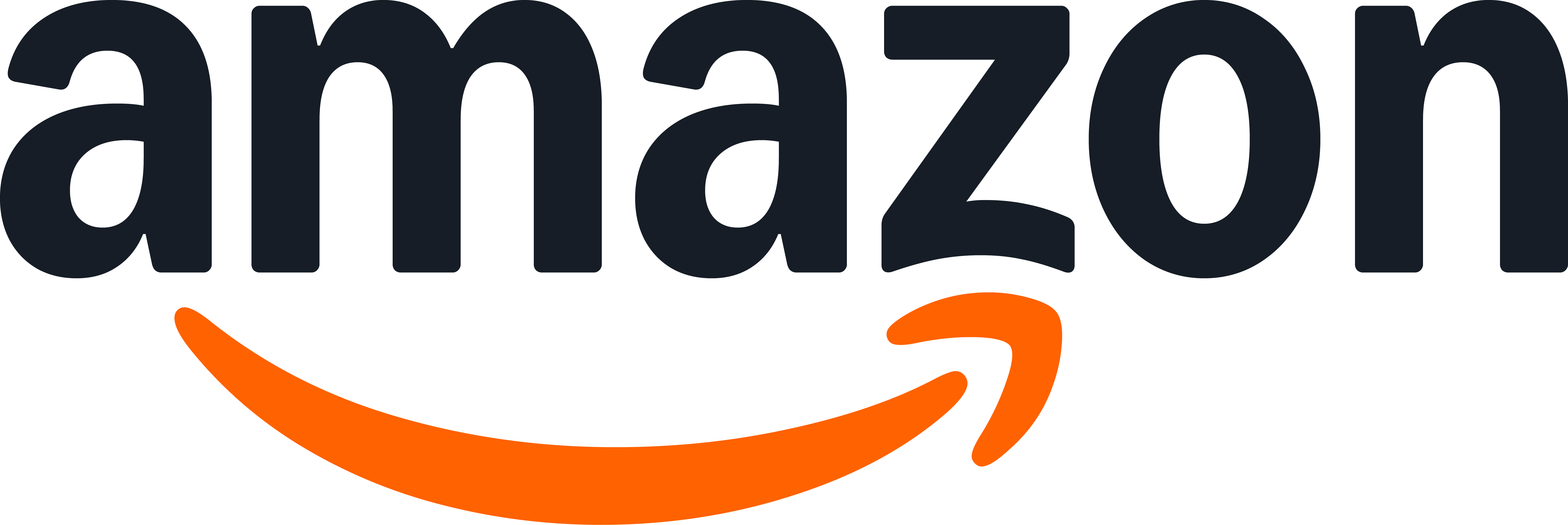}}
  \\
  \And
  Behrooz Omidvar-Tehrani
  \textsuperscript{ \includegraphics[height=1.8em]{logos/AWS_logo_RGB_1c_Gray850-2.png}}
  \\
  \And
  Jun Huan
  \textsuperscript{ \includegraphics[height=1.8em]{logos/AWS_logo_RGB_1c_Gray850-2.png}}
  \\
  \And
  Anoop Deoras
  \textsuperscript{ \includegraphics[height=1.8em]{logos/AWS_logo_RGB_1c_Gray850-2.png}}
  \\
  \And
  Stefano Soatto
  \textsuperscript{ \includegraphics[height=1.8em]{logos/AWS_logo_RGB_1c_Gray850-2.png}}
  \\
  \AND
  \\
  \vspace{-1cm}
  \includegraphics[height=0.7em]{logos/mit_logo_std_cmyk_mit-red_cropped.jpg} Massachusetts Institute of Technology
  \quad
  \includegraphics[height=0.8em]{logos/AWS_logo_RGB_1c_Gray850-2.png} AWS AI
  \quad
  \includegraphics[height=0.8em]{logos/Amazon_Logo_Squid_Ink_Smile_Orange.png} Amazon
}

\begin{document}

\maketitle

\begin{abstract}
Reinforcement Learning with Verifiable Rewards (RLVR) has become a standard recipe for post-training LLMs on reasoning tasks, with Group Relative Policy Optimization (GRPO) emerging as a leading approach. However, GRPO and its variants are inherently single-turn: they optimize from terminal rewards on isolated prompt-response pairs, leaving them poorly suited to agentic settings where models must iteratively refine solutions in response to environmental feedback. We introduce \method, a multi-turn extension of
GRPO for self-correcting code generation. \method~constructs
feedback-conditioned rollout trees in which failed candidate solutions are paired with executor feedback and expanded into subsequent turns, and propagates rewards backward through the tree so that later successful refinements credit earlier attempts that surfaced
informative feedback. We study two propagation strategies, Max Reward (\textsc{MaRS}) and Mean Reward (\textsc{MeRS}), and introduce post-rollout pruning mechanisms that reduce multi-turn optimization cost. Across three code generation benchmarks (HumanEval, MBPP, LiveCodeBench-v6) and two model families (Qwen3-1.7B/4B, OLMo-2-7B), \method~delivers up to \textbf{6\%} absolute pass@1 gains over the strongest prior multi-turn execution-feedback methods. Gains are largest on the Medium/Hard subset (\textbf{+4.38/+4.20} at Iter-$5$), where iterative self-correction matters more.
\end{abstract}

\input{section/introduction}

\input{section/background}

\input{section/proposed_approach}

\input{section/experiments}

\input{section/related_work}

\input{section/conclusion}

\newpage

\bibliographystyle{abbrv}

\bibliography{section/references}

\newpage

\include{section/appendix}



\end{document}

%% file: section/introduction.tex
\section{Introduction}
\label{sec:introduction}

\begin{qoutebox}

\textit{``The road to wisdom? Well, it’s plain and simple to express: err and err and err again, but less and less and less.'' \hfill ---\textit{Piet Hein}}
\end{qoutebox}

\definecolor{segA}{HTML}{4E79A7} 
\definecolor{segA_dark}{HTML}{000000} 

\definecolor{segB}{HTML}{59A14F} 
\definecolor{segB_dark}{HTML}{000000} 

\begin{figure}[h!]
    \centering
    \includegraphics[width=1\linewidth]{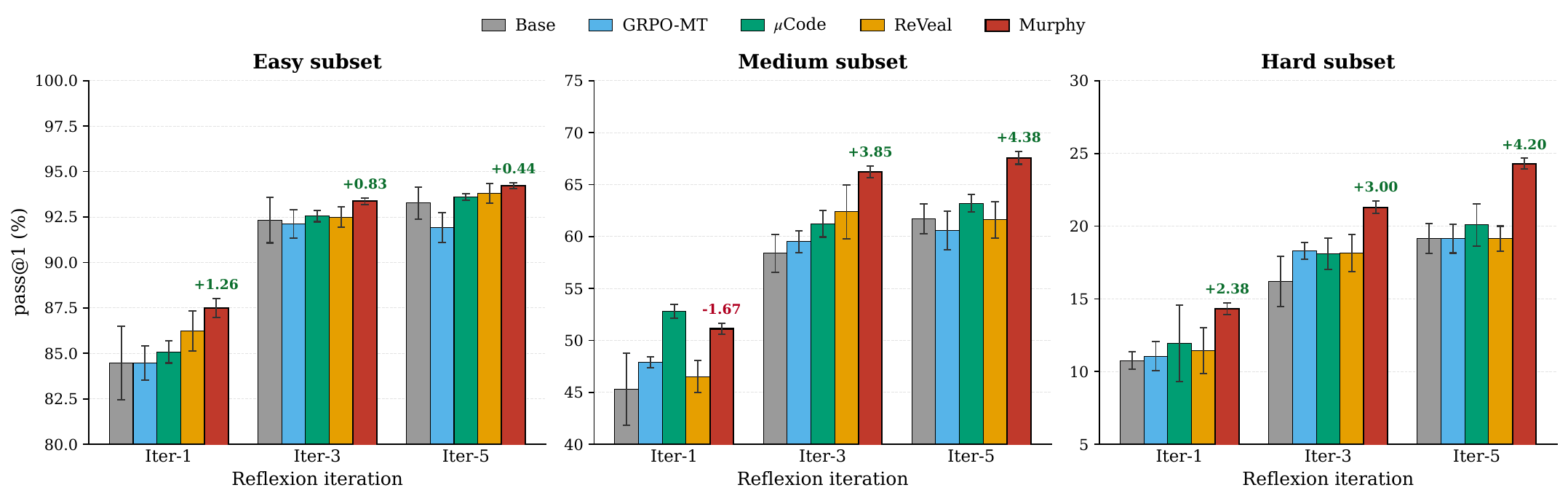}
    \caption{\textbf{Self-correction performance on LiveCodeBench-v6.}
    Qwen3-4B with \method~vs. four execution-feedback baselines (Base,
    GRPO-MT, $\mu$\textsc{Code}, \textsc{ReVeal}) on the Easy, Medium, and
    Hard subsets. Base, GRPO-MT, $\mu$\textsc{Code}, and \method~are evaluated
    under the Reflexion framework~\citep{reflexion}; \textsc{ReVeal} is
    evaluated using its official generation-verification scaffold, which
    uses a method-specific prompt and parser. All methods are compared at
    1, 3, and 5 inference turns. Bars show the mean and stdev across 3 runs.
    \method shows large gains on the Medium/Hard subset (\textbf{+4.38/+4.20} at
    Iter-$5$), where self-correction matters more.}
    \label{fig:livecodebench}
\end{figure}

Large language models (LLMs) are increasingly deployed as code-generation agents that interact with their environment rather than producing a single static response~\citep{sweagent, wang2025openhands}. In a typical execution-feedback scaffold~\citep{reflexion, lats, dot}, the model proposes code, the code is executed against unit tests, and failures are returned as error messages, stack traces, or failing test cases. The model is then re-prompted with the original task, its previous attempt, and the feedback, and this process repeats until the solution is accepted or a turn budget is exhausted. Such scaffolds substantially improve inference-time self-correction, but they leave the underlying model unchanged: feedback is consumed as test-time context only~\citep{reflexion, e3}.

Recent works attempt to close this gap by training models to use execution feedback. RLEF~\citep{rlef} optimizes multi-turn code generation with PPO~\citep{ppo} against execution-based rewards. $\mu$\textsc{Code}~\citep{multiturncodegen} recasts multi-turn correction as expert-iteration imitation learning with a learned verifier for relabeling and selection. \textsc{ReVeal}~\citep{reveal} interleaves generation and verification within a single policy, using turn-aware optimization to improve self-verification. These methods establish the value of execution feedback, but each relies on PPO-style value estimation, learned verifiers, or explicit verification policies. They leave open a simpler question: \textbf{can the standard GRPO}~\citep{grpo} \textbf{post-training recipe itself be made feedback-aware in multi-turn environments?}

A natural baseline is GRPO-MT, which extends GRPO by treating each feedback-conditioned interaction as a trajectory. The model samples candidates for the original prompt; passing rollouts terminate, while failed ones are paired with executor feedback and expanded for additional turns until solved or the turn budget is exhausted. Each terminated rollout's final execution reward serves as its trajectory-level outcome for the GRPO advantage, broadcast across policy generated tokens along that rollout while executor feedback and prior-solution context are masked from the loss. Although GRPO-MT exposes the model to execution feedback, its credit assignment is coarse: all generations along a terminated trajectory inherit the same terminal signal, regardless of whether an intermediate attempt produced useful feedback, a misleading direction, or contributed little to the eventual fix. This matters in code generation, where failed attempts are not uniformly bad: an incorrect solution may reveal the precise failing case or runtime behavior that enables a later refinement. GRPO-MT can reinforce successful trajectories but cannot distinguish informative failures from uninformative ones when they share a terminal outcome.

Prior inference-time agent frameworks~\citep{reflexion, lats, dot} show that previous failed attempts and environment feedback provide useful signals for self-correction. In code generation, a failed program may expose an edge case, runtime error, or test failure that enables a later refinement. Thus, failed attempts should not be treated as uniformly negative; their value depends on whether the feedback they induce supports future correction.

\method~turns this observation into a training-time objective for multi-turn GRPO. It constructs feedback-conditioned rollout trees in which successful candidates terminate and failed candidates are expanded by re-prompting on the original problem, prior output, and executor feedback. Once the rollout is complete, \method~propagates rewards backward through the tree, allowing later successful refinements to credit earlier attempts. We instantiate this idea with Max Reward (\textsc{MaRS}), which backs up the best descendant outcome, and Mean Reward (\textsc{MeRS}), which backs up a discounted average future reward, then apply a local GRPO update using the resulting credit-adjusted rewards. This design isolates the algorithmic question from confounding system choices. Unlike RLEF, \method~requires no PPO critic; unlike $\mu$\textsc{Code}, no learned verifier or imitation reduction; unlike \textsc{ReVeal}, no separate verification behavior or test synthesis. \method~is a minimal, architecture-preserving extension of GRPO to feedback-conditioned, multi-turn code generation.

\begin{mainbox}

\begin{enumerate}
\item We introduce \method, a multi-turn GRPO algorithm that trains code LLMs to use execution feedback through feedback-conditioned rollouts and structured temporal credit assignment. (\autoref{sec:murphy})
\item We propose \textsc{MaRS} and \textsc{MeRS}, two reward-propagation strategies that assign credit from later refinements to earlier attempts in the rollout tree.
\item We design pruning strategies that make multi-turn GRPO training tractable by retaining informative trajectories while reducing optimization cost. (\autoref{subsec:murphy_with_pruning})
\item We evaluate \method~on three code-generation benchmarks across two model families (OLMo, Qwen) and sizes (1.7B–7B), achieving up to \textbf{6\%} absolute pass@1 gains over prior multi-turn execution-feedback methods. (\autoref{sec:experiments})
\end{enumerate}
\end{mainbox}

%% file: section/background.tex
\section{Background: GRPO}
\label{sec:background}

Group Relative Policy Optimization (GRPO; \citep{grpo}) is a reinforcement-learning objective for LLM fine-tuning that avoids learning a separate value function. For each input prompt $q$, the old policy $\pi_{\theta_{\mathrm{old}}}$ samples a group of $G$ candidate responses, $
    \mathcal{C}(q)=\{u_1,\ldots,u_G\} $. Each response $u \in \mathcal{C}(q)$ receives a scalar reward $r(u)$. GRPO computes advantages by normalizing rewards within the response group:
\begin{align}
    \hat A^{\mathrm{GRPO}}_{q}(u)
    =
    \frac{
        r(u) -
        \mathrm{mean}_{w \in \mathcal{C}(q)} r(w)
    }{
        \mathrm{std}_{w \in \mathcal{C}(q)} r(w)
    },
    \qquad u \in \mathcal{C}(q)
    \label{eq:grpo_advantage}
\end{align}
This group-relative baseline replaces the learned critic used in PPO. Let $u_t$ denote the $t$-th token of response $u$, and let $u_{<t}$ be its prefix. The token-level importance ratio is $R_{\theta}(q,u,t)
    =
    \frac{
        \pi_\theta(u_t \mid q,u_{<t})
    }{
        \pi_{\theta_{\mathrm{old}}}(u_t \mid q,u_{<t})
    }$. The GRPO clipped objective for a response group is then
\begin{align}
    \mathcal{J}_{\mathrm{GRPO}}\left(\theta; q, \mathcal{C}(q), \hat A_q\right)
    &=
    \mathbb{E}_{u \in \mathcal{C}(q)}
    \bigg[
    \frac{1}{|u|}
    \sum_{t=1}^{|u|}
    \min\!\Big(
        R_{\theta}(q,u,t)\hat A_q(u), \notag \mathrm{clip}(R_{\theta}(q,u,t)\\
    &\qquad\qquad\quad
        ,1-\epsilon,1+\epsilon)\hat A_q(u)
    \Big)
    \bigg]
    -\, \beta\, D_{\mathrm{KL}}\!\left(\pi_\theta \,\|\, \pi_{\mathrm{ref}}\right)
    \raisetag{2.5\baselineskip}
    \label{eq:grpo_objective}
\end{align}
where $\pi_{\mathrm{ref}}$ is a fixed reference policy and $\beta$ controls the KL penalty. In standard single-turn GRPO, $\hat A_q=\hat A^{\mathrm{GRPO}}_q$ is computed from the immediate rewards of responses sampled for the same prompt. \method~keeps this group-relative update structure but changes how the rewards used in the advantage are assigned in a multi-turn rollout tree as described in \autoref{sec:murphy}.

%% file: section/proposed_approach.tex
\section{Proposed Method: \method}
\label{sec:murphy}

\begin{figure*}[htbp]
  \centering
  \includegraphics[width=1.0\textwidth]{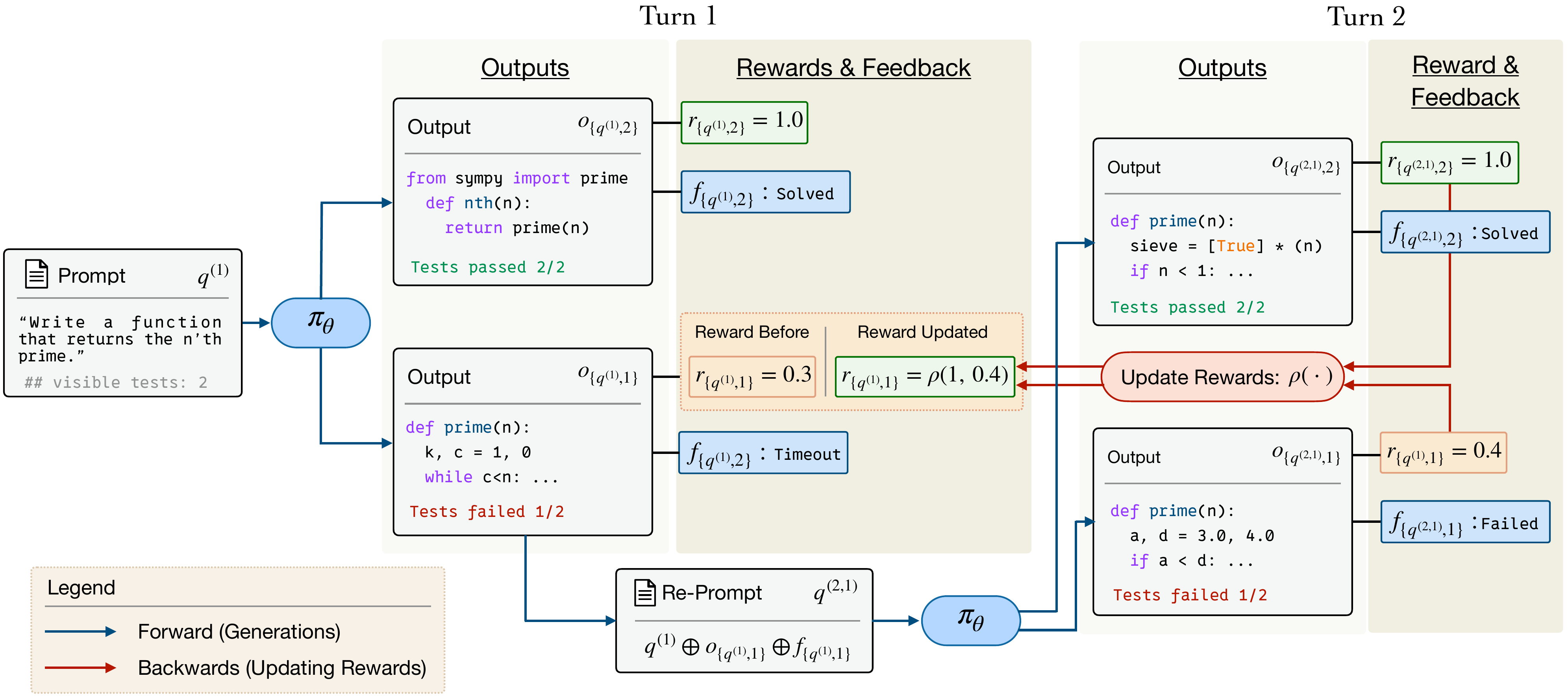}
  \caption{\textbf{Overview of \method.} Given an input prompt $q_{(1)}$, the model samples multiple candidate solutions. Each output is executed to obtain a scalar reward $r$ and executor feedback $f$. Solved outputs terminate, while failed outputs are converted into feedback-conditioned prompts by concatenating the original prompt, the failed output, and its feedback. The model then samples refinements for each failed output. After the rollout tree is generated, rewards from later turns are propagated backward through a credit-assignment rule $\rho(\cdot)$, allowing successful refinements to credit earlier failed attempts that produced useful feedback.}
  \label{fig:main_figure}
\end{figure*}

\method~extends GRPO to feedback-rich, multi-turn code generation. Standard GRPO optimizes one response group per prompt using rewards from that same generation step. This is insufficient for agentic code generation, where the model may execute code, observe failures, and refine its solution over multiple attempts. In this setting, an initially incorrect solution can still be valuable if it exposes useful executor feedback, such as a failing edge case or runtime error, that enables a later correction. \method~addresses this mismatch by constructing feedback-conditioned rollout trees and assigning credit retrospectively from later successful refinements to earlier attempts.

\paragraph{Feedback-conditioned rollout tree.}
For each input prompt, \method~first samples a group of candidate programs. Each candidate is executed against the task test suite, producing two signals: a scalar reward, such as the proportion of tests passed, and qualitative executor feedback, such as failing test cases, stack traces, compiler errors, or timeout messages. Candidates that achieve the maximum reward are treated as solved and become terminal nodes: they are kept for optimization but are not expanded in later turns.

Failed candidates are expanded in the next turn. To do so, \method~forms a new prompt by concatenating the original task prompt, the failed output, and the corresponding executor feedback. The policy is then re-invoked on this feedback-conditioned prompt to generate a new group of candidate refinements. This expansion is local to each failed node: different failures produce different feedback-conditioned prompts and therefore induce different refinement groups. The process repeats for a fixed number of turns or until no failed nodes remain. The result is a rollout tree in which each node is a generated program and each edge corresponds to one refinement step conditioned on executor feedback from the parent node.

This tree structure preserves the causal relationship between an attempted solution, the feedback it produces, and the refinements that follow. In \autoref{fig:main_figure}, the first turn produces two outputs. The solved output terminates immediately. The failed output receives feedback, is converted into a new prompt, and is expanded into two second-turn refinements. Thus, later outputs are not independent samples from the original prompt; they are feedback-conditioned descendants of specific earlier attempts.

\paragraph{Retrospective credit assignment.}
After the rollout tree is generated and all candidates are evaluated, \method~propagates rewards backward using a bottom-up recursion. Starting from the leaves, each node passes reward information to its parent; the parent is then updated using its own immediate execution reward and the adjusted rewards of its children. This differs from single-turn GRPO, where a response is credited only by its own reward within the original response group. In \method, a failed attempt can receive positive credit if its feedback leads to a later successful refinement.

\paragraph{Credit-adjusted rewards.}
Let $\mathcal{T}$ denote the rollout tree for a prompt, where each node $v \in \mathcal{T}$ corresponds to a generated program. Each node has an immediate execution reward $r(v)$ and, if expanded, a child response group $\mathcal{C}(v)$ generated by re-prompting on its feedback. Solved nodes are terminal and have $\mathcal{C}(v)=\emptyset$. \method~computes credit-adjusted rewards $\tilde r(v)$ by a backward recursion over $\mathcal{T}$. For leaf nodes, $\tilde r(v)=r(v)$. For rest nodes, we consider 2 credit-assignment rules:\looseness=-1
\begin{align}
    \textsc{MaRS}: \quad
    \tilde r(v)
    &=
    \max\left(r(v), \max_{u \in \mathcal{C}(v)} \tilde r(u)\right),
    \label{eq:mars}
    \\
    \textsc{MeRS}: \quad
    \tilde r(v)
    &=
    \frac{
        r(v) + \gamma \cdot
        \mathrm{mean}_{u \in \mathcal{C}(v)} \tilde r(u)
    }{
        Z(v)
    },
    \label{eq:mers}
\end{align}

where $\gamma \in [0,1]$ is a discount factor and $Z(v)$ is a normalization term that counts the reward components included in the backup. This keeps adjusted rewards comparable across nodes at different depths. In our two-turn setting, $Z(v)=2$ for unsolved internal nodes, corresponding to the node's immediate reward and the propagated reward from its children, while terminal nodes retain $Z(v)=1$ and $\tilde r(v)=r(v)$. \textsc{MaRS} performs an optimistic backup: it propagates the best achievable refinement outcome upward through the tree, similar in spirit to max-backup variants of Monte Carlo Tree Search~\citep{khandelwal2016complex}. Thus, if one refinement of a failed first-turn solution solves the task, \textsc{MaRS} credits that success back to the first-turn node. \textsc{MeRS} instead propagates an average future-refinement value, analogous to a Bellman-style backup over sampled refinement nodes. Unlike value-based RL, however, \method~performs this backup directly over generated refinements rather than learning a separate value function.

\paragraph{Node-wise GRPO objective.}
After credit assignment, \method~applies GRPO locally to each response group in the rollout tree. Let $\mathcal{Q}(\mathcal{T})$ denote the set of prompts and feedback-conditioned prompts expanded in the tree. For any $q \in \mathcal{Q}(\mathcal{T})$, let $\mathcal{C}(q)$ be its response group. Instead of computing advantages from immediate rewards as in \autoref{eq:grpo_advantage}, \method~normalizes credit-adjusted rewards within the group:
\begin{align}
    \hat A^{\method}_{q}(u)
    =
    \frac{
        \tilde r(u) -
        \mathrm{mean}_{w \in \mathcal{C}(q)} \tilde r(w)
    }{
        \mathrm{std}_{w \in \mathcal{C}(q)} \tilde r(w)
    },
    \qquad u \in \mathcal{C}(q).
    \label{eq:murphy_advantage}
\end{align}
The \method objective is the sum of GRPO losses over the expanded response groups:
\begin{align}
    \mathcal{J}_{\method}(\theta)
    =
    \mathbb{E}_{\mathcal{T} \sim \pi_{\theta_{\mathrm{old}}}}
    \left[
        \sum_{q \in \mathcal{Q}(\mathcal{T})}
        \mathcal{J}_{\mathrm{GRPO}}
        \left(
            \theta;
            q,
            \mathcal{C}(q),
            \hat A^{\method}_{q}
        \right)
    \right]
    \label{eq:murphy_objective}
\end{align}
Thus, \method~does not introduce a critic or a new policy-gradient estimator. It applies the same GRPO update from \autoref{eq:grpo_objective} at each expanded response group in the rollout tree, while replacing immediate rewards with retrospectively adjusted rewards. The policy update remains token-level, but the scalar learning signal attached to each generation now reflects both its immediate execution result and its contribution to later feedback-conditioned refinements. We provide the full rollout-tree indexing, recursive credit-assignment details, and implementation details in \autoref{app:murphy_formalism}. 

\subsection{Pruning Strategies in \method}
\label{subsec:murphy_with_pruning}

The feedback-conditioned rollout tree in \method~can grow quickly with the number of turns. By default, \method~uses a fixed generation budget $G$ at every turn. If most generations fail and continue to be expanded, the number of generated programs can grow as $G^S$ for a turn budget $S$, in the worst case. This increases memory use and slows optimization, since each retained sequence may contribute to advantage computation and gradient updates. To make multi-turn training tractable, we introduce pruning strategies that retain a smaller subtree before retrospective credit assignment.

Pruning is applied after rollouts are generated and rewards are computed, but before rewards are propagated backward. This order is important: the rollout procedure still collects feedback-conditioned generations and their rewards, but only the retained subtree contributes to credit assignment and optimization. Let $\mathcal{T}' \subseteq \mathcal{T}$ denote the retained subtree after pruning $(\mathcal{T}' = \mathrm{Prune}(\mathcal{T}))$. Pruning changes both the children used in the backward credit recursion and the response groups included in the node-wise GRPO objective:
\begin{align}
    \mathcal{J}_{\method}^{\mathrm{pruned}}(\theta)
    &=
    \mathbb{E}_{\mathcal{T} \sim \pi_{\theta_{\mathrm{old}}}}
    \left[
        \sum_{q \in \mathcal{Q}(\mathcal{T}')}
        \mathcal{J}_{\mathrm{GRPO}}
        \left(
            \theta;
            q,
            \mathcal{C}_{\mathcal{T}'}(q),
            \hat A^{\method}_{q}
        \right)
    \right],
    \label{eq:murphy_pruned_objective}
\end{align}
where $\mathcal{C}_{\mathcal{T}'}(q)$ denotes the retained response group for prompt $q$ in the pruned tree. In this sense, pruning is a compute-aware filter on the rollout tree: it does not change the rollout mechanism or the GRPO update, but determines which generated trajectories are retained for reward propagation and gradient updates.

We consider two pruning strategies with different budget granularities. \textsc{IntraP} uses a \emph{within-group trajectory budget}, retaining a fixed number of trajectories inside each response group. \textsc{InterP} uses an \emph{across-group budget}, retaining a fixed number of refinement groups among those induced by failed parent nodes. We use $\mathrm{TopB}^{b}_{x \in \mathcal{X}}(s(x))$ to denote the subset of $b$ elements from $\mathcal{X}$ with the largest scores $s(x)$.

\paragraph{Intra-Group Pruning (\textsc{IntraP}).}
\textsc{IntraP} prunes individual trajectories within each response group. Consider a prompt or feedback-conditioned prompt $q$ with a group of generated children $\mathcal{C}(q)$. Given a within-group trajectory budget $b_{\mathrm{traj}}$, \textsc{IntraP} retains a subset $\mathcal{C}'(q) \subseteq \mathcal{C}(q)$ with $|\mathcal{C}'(q)| = b_{\mathrm{traj}}$ and discards the remaining children together with their descendants. The retained trajectories are selected according to their contribution to within-group reward variance: $\mathcal{C}'(q)
    =
    \mathrm{TopB}_{u \in \mathcal{C}(q)}^{b_{\mathrm{traj}}}
    \left(
        \Delta_{\mathrm{Var}}(u;\mathcal{C}(q))
    \right),$ where $\Delta_{\mathrm{Var}}(u;\mathcal{C}(q))$ measures how much trajectory $u$ contributes to the reward variance of the group. This strategy is inspired by \cite{pods}, who show that retaining high-variance trajectories within a response group can reduce optimization cost while preserving the learning signal needed for group-relative policy optimization. We extend this idea to the multi-turn setting by applying it recursively over the rollout tree. Since GRPO computes advantages from relative reward differences within a group, preserving reward diversity helps retain informative comparisons while reducing the number of sequences used for optimization. If retained samples have nearly identical rewards, the normalized advantages provide little useful preference signal; \textsc{IntraP} therefore keeps trajectories that maintain reward contrast within each group.

\paragraph{Inter-Group Pruning (\textsc{InterP}).}
\textsc{InterP} prunes at the level of entire refinement groups. Each failed node at turn $s$ creates a feedback-conditioned prompt at turn $s+1$, and that prompt produces a group of child refinements. Instead of selecting individual trajectories within every group, \textsc{InterP} selects which refinement groups to retain. The intuition is that not all feedback contexts are equally useful for optimization. Some failed attempts lead to refinement groups where all children behave similarly: they may all solve the task, or they may all fail. Such groups provide limited relative learning signal because there is little contrast among candidates generated from the same feedback-conditioned prompt. Other failed attempts lead to mixed refinement groups, where some children improve substantially while others still fail. These groups are more informative because they reveal which refinements should be preferred under the same feedback context. Based on this intuition, \textsc{InterP} ranks refinement groups by their within-group reward variance. Let $\mathcal{G}_s$ denote the set of child groups induced by failed nodes at turn $s$. Given an across-group budget $b_{\mathrm{grp}}$, \textsc{InterP} retains the $b_{\mathrm{grp}}$ groups with the largest within-group reward variance: $
    \mathcal{G}'_s
    =
    \mathrm{TopB}_{g \in \mathcal{G}_s}^{b_{\mathrm{grp}}}
    \left(
        \mathrm{Var}\big(\{r(u): u \in g\}\big)
    \right)$. The remaining groups, along with all of their descendants, are discarded.

This pruning rule is aligned with the group-relative nature of GRPO. Since advantages are computed by comparing rewards within a group, high-variance groups provide stronger preference information than groups whose rewards are nearly uniform. In contrast to \textsc{IntraP}, which keeps informative trajectories within each group, \textsc{InterP} allocates optimization budget across feedback-conditioned prompts by keeping the refinement groups that appear most informative for learning. In \autoref{app:efficiency_gains}, we provide a discussion on the practical cost of rollout tree generation.

%% file: section/experiments.tex
\section{Experiments}
\label{sec:experiments}

In this section, we first provide an overview of metrics and evaluation protocol used in our experiments, followed by detailed results, including ablation studies. We provide implementation details including inference hyper-parameters in~\autoref{app:implementation_details}.

{\textbf{Metrics and Evaluation Protocol.}}
Reflexion~\citep{reflexion} is a widely adopted purely inference-time
multi-turn framework: the model parameters are not updated, but the model
iteratively refines its solution by conditioning on prior attempts,
execution feedback from visible test cases, and self-generated reflections
(see~\autoref{app:reflexion} for details). To assess refinement and
self-correction, we integrate all models into the same Reflexion scaffold
and report \textit{pass@1} under $K$-iteration evaluation. We denote these
settings as Iter-$K$: Iter-1 corresponds to standard input-output prompting
without feedback, while at iteration $k>1$ the model revises its solution
using a sliding-window history consisting of the previous output and its
corresponding visible-test feedback. Unless otherwise specified, we use
Iter-3 as the default multi-turn evaluation setting. The agent terminates
early once all visible tests pass or when the maximum iteration budget is
reached. Final solutions are evaluated on hidden test cases, and the
resulting \textit{pass@1} is reported. Each experiment is repeated three
times, and we report the mean and standard deviation of \textit{pass@1}
across runs.

\begin{figure}[htbp!]
    \centering
    \includegraphics[width=0.9\linewidth]{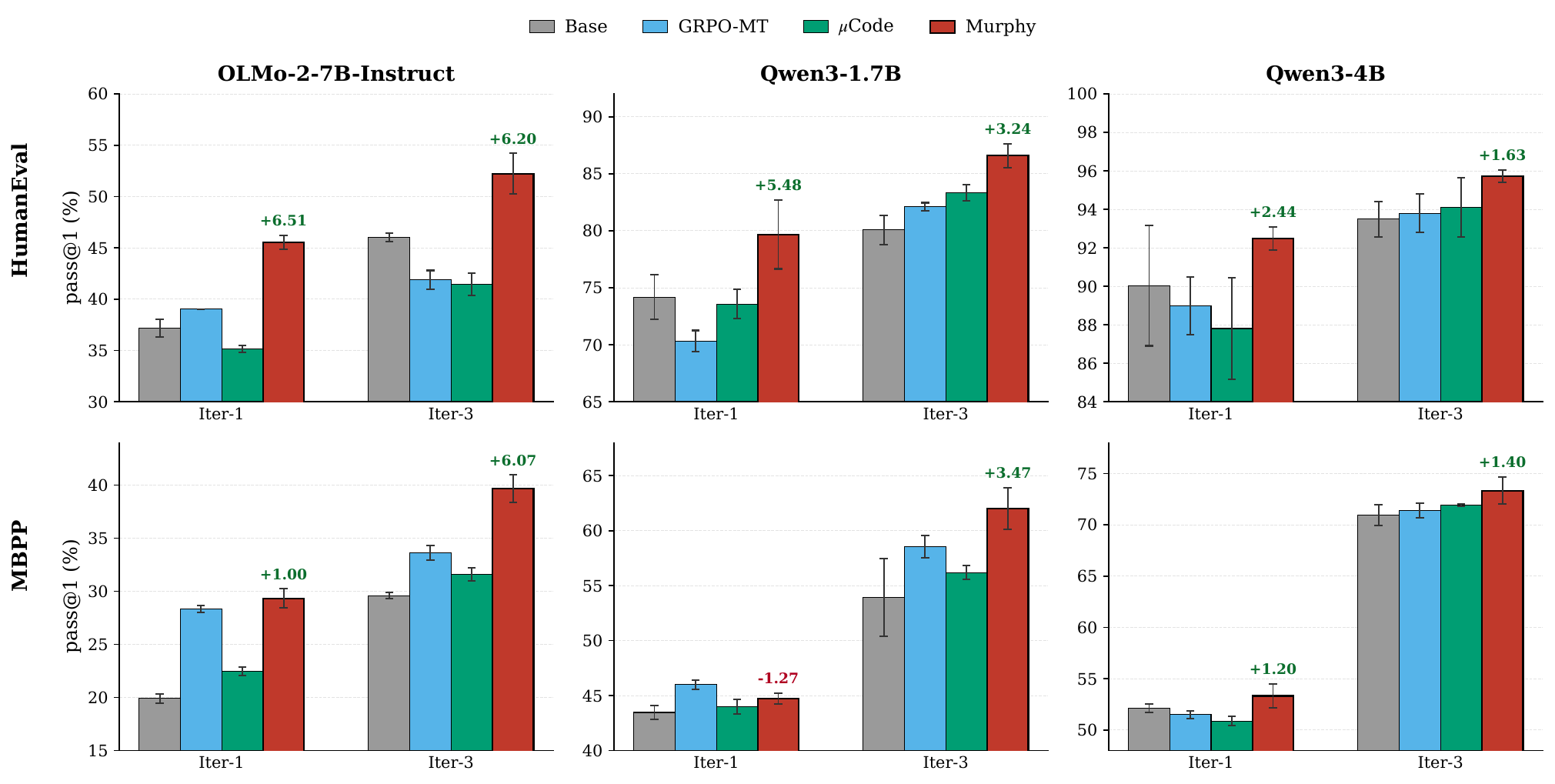}
    \caption{\textbf{HumanEval and MBPP pass@1 across Reflexion iterations
    for three base models.} Rows correspond to benchmarks (HumanEval, MBPP)
    and columns to base models (OLMo-2-1124-7B-Instruct, Qwen3-1.7B,
    Qwen3-4B). Within each cell, we compare \textsc{Murphy} (ours) against
    the Base model, the multi-turn baseline GRPO-MT, and the
    execution-feedback method $\mu$Code~\citep{multiturncodegen}. Annotations
    above \textsc{Murphy}'s bars report the absolute percentage-point
    difference between \textsc{Murphy} and the strongest non-\textsc{Murphy}
    baseline within the same benchmark, model, and Reflexion iteration
    (green: improvement, red: regression). \textsc{Murphy} offers superior performance in Iter-3 settings, where feedback-conditioned self-correction is directly evaluated. Detailed results in~\autoref{tab:humaneval_mbpp}.}
    \label{fig:humaneval_mbpp}
\end{figure}

\paragraph{HumanEval and MBPP.}
We evaluate OLMo-2-1124-7B-Instruct~\citep{olmo2} and Qwen3
(1.7B, 4B)~\citep{Qwen3}, each fine-tuned on 1{,}000 samples randomly
drawn from KodCode~\citep{xu2025kodcode} and evaluated on
HumanEval~\citep{chen2021humaneval} and MBPP~\citep{austin2021mbpp}
using the evaluation protocol described above. KodCode is
chosen for its minimal overlap with evaluation benchmarks\footnote{Refer
to Section~3.2 of KodCode~\citep{xu2025kodcode} for details on
contamination analysis.}. We report \textit{pass@1} under single-iteration
(Iter-1) and three-iteration (Iter-3) Reflexion settings in
\autoref{fig:humaneval_mbpp}; in all experiments, \method~uses
\textsc{MaRS} reward propagation.

Iter-1 evaluates one-shot generation and does not exercise
feedback-conditioned refinement, so we treat it primarily as a sanity
check that multi-turn training does not substantially degrade direct
generation. In this setting, \method~is competitive with the strongest
baseline with one small regression on Qwen3-1.7B MBPP
(\textbf{-1.27 pp}). The primary evaluation for self-correction is
Iter-3, where models can condition on execution feedback and revise
their previous attempts. Under this setting, \method~outperforms all
baselines, with gains up to~\textbf{+6.20 pp} over the strongest comparator. The largest improvements
occur on OLMo-2-1124-7B-Instruct (\textbf{+6.20 pp} on HumanEval and
\textbf{+6.07 pp} on MBPP), suggesting that weaker models benefit more
from structured multi-turn credit assignment. Overall, these results
indicate that retrospective reward propagation primarily improves
feedback-conditioned refinement while largely preserving one-shot
generation performance.

\paragraph{LiveCodeBench-v6.}
We additionally include \textsc{ReVeal}~\citep{reveal}, a concurrent
multi-turn RL framework that evolves code generation through
self-verification and tool-based evaluation, in our most rigorous
setting: the contamination-resistant LiveCodeBench-v6 benchmark with
Qwen3-4B (\autoref{fig:livecodebench}). Base, GRPO-MT,
$\mu$\textsc{Code}, and \method~are evaluated under the same Reflexion
scaffold. For \textsc{ReVeal}, we instead use its official repository,
hyperparameters, and native multi-turn generation-verification scaffold
with 3 training turns. This gives \textsc{ReVeal} its intended evaluation
setting, since its inference procedure relies on a method-specific
generation prompt, output format, and parser. Although the prompt and
parser differ from Reflexion, the underlying scaffolds are conceptually
similar: both condition on prior incorrect responses and environment
feedback over multiple turns to refine subsequent attempts. All methods
are compared at 1, 3, and 5 inference turns. \method~outperforms both
$\mu$\textsc{Code} and \textsc{ReVeal} across all difficulty subsets,
with the largest gain (+4.20 pp) on the Hard subset at Iter-5, where
self-correction matters most. Detailed results, including scaffold
robustness analyses, are available in~\ref{app:lcb-v6} (\method’s gains remain consistent across multiple scaffolds).
\looseness=-1

\paragraph{Scope of \textsc{ReVeal} comparison.} 
\textsc{ReVeal}'s co-trained self-verifier substantially increases per-experiment 
training cost: \textasciitilde{}23 hours end-to-end for Qwen3-4B versus 
\textasciitilde{}11 hours for \method~on matched hardware 
(8$\times$H100s). We therefore include ReVeal as a baseline only in our 
most rigorous setting (LCB-v6 with Qwen3-4B, \autoref{fig:livecodebench}), 
where contamination resistance and difficulty stratification make the 
comparison most informative. Extending \textsc{ReVeal} across all 
model–benchmark cells in \autoref{fig:humaneval_mbpp} was prohibitive 
under our compute budget, mirroring our handling of RLEF~\citep{rlef}.

\textbf{Additional Experiments.} Further analysis is provided in the appendices: effect of training data size (App.\ref{subsec:effect_of_training_data_size}), and training beyond two turns (App.\ref{app:3_stage}). App.~\ref{app:scaffold_sensitivity} confirms that \method's gains generalize across different iterative scaffolds.\looseness=-1

\subsection{Reward Propagation Ablation: \textsc{MaRS} vs.\ \textsc{MeRS}}
We ablate the two reward propagation strategies, \textsc{MaRS} and 
\textsc{MeRS}, introduced in~\autoref{subsec:murphy_with_pruning}, training Qwen3-1.7B 
and OLMo-2-1124-7B-Instruct on 1K KodCode subset dataset under each 
strategy (\autoref{tab:ablation_max_vs_value}). \textsc{MaRS} 
offers competitive or superior performance over \textsc{MeRS} in iter-3 setting independent of the discount factor $\gamma$. 
The gap stems from how each strategy handles non-binary rewards: 
\textsc{MeRS} averages over child rewards, diluting the signal when 
few refinements score highly, whereas \textsc{MaRS} propagates the 
strongest outcome, letting rare high-reward trajectories dominate the 
update. This advantage is most pronounced under sparse rewards and 
diminishes as rewards become binary.\looseness=-1

\newcommand{\comprow}{\rowcolor{gray!4}}

\newcommand{\deltap}[2]{\cellcolor{green!#2}{\textbf{+}#1}}
\newcommand{\deltan}[1]{\cellcolor{gray!10}{\textbf{--}#1}}

\begin{table*}[t]
\centering
\caption{\textsc{MaRS} vs.\ \textsc{MeRS} reward propagation. \textsc{MaRS} is highlighted. $\Delta_3$ is relative to \textsc{MeRS} ($\gamma=1$) Iter-3 within each model block.}
\label{tab:ablation_max_vs_value}
\scriptsize
\setlength{\tabcolsep}{3.5pt}
\renewcommand{\arraystretch}{0.92}
\resizebox{0.95\textwidth}{!}{%
\begin{tabular}{l ccc ccc}
\toprule
\textbf{Model / Strategy}
& \multicolumn{3}{c}{\textbf{HumanEval}} 
& \multicolumn{3}{c}{\textbf{MBPP}} \\
\cmidrule(lr){2-4} \cmidrule(lr){5-7}
& \textbf{Iter-1} & \textbf{Iter-3} & \textbf{$\Delta_3$}
& \textbf{Iter-1} & \textbf{Iter-3} & \textbf{$\Delta_3$} \\
\midrule

\multicolumn{7}{l}{\textbf{Qwen3-1.7B}} \\
\oursrow
\quad \textsc{MaRS}
& \textbf{79.67 $\pm$ 3.01} & \textbf{86.58 $\pm$ 1.06} & \deltap{1.01}{10}
& 44.73 $\pm$ 0.50 & \textbf{62.00 $\pm$ 1.91} & \deltap{1.07}{10} \\

\comprow
\quad \textsc{MeRS} ($\gamma=0.9$)
& \underline{78.66 $\pm$ 2.11} & 84.76 $\pm$ 1.22 & \deltan{0.81}
& \textbf{46.53 $\pm$ 0.81} & 60.53 $\pm$ 1.79 & \deltan{0.40} \\

\comprow
\quad \textsc{MeRS} ($\gamma=1$)
& 78.46 $\pm$ 0.70 & \underline{85.57 $\pm$ 0.35} & \deltaz
& \underline{45.07 $\pm$ 1.10} & \underline{60.93 $\pm$ 1.29} & \deltaz \\

\midrule
\multicolumn{7}{l}{\textbf{OLMo-2-1124-7B-Instruct}} \\
\oursrow
\quad \textsc{MaRS}
& \textbf{45.53 $\pm$ 0.70} & \textbf{52.24 $\pm$ 1.96} & \deltap{10.37}{80}
& \textbf{29.33 $\pm$ 0.90} & \textbf{39.67 $\pm$ 1.29} & \deltap{5.27}{45} \\

\comprow
\quad \textsc{MeRS} ($\gamma=1$)
& \underline{37.40 $\pm$ 0.35} & \underline{41.87 $\pm$ 1.27} & \deltaz
& \underline{27.87 $\pm$ 0.12} & \underline{34.40 $\pm$ 2.40} & \deltaz \\

\bottomrule
\end{tabular}%
}
\vspace{-0.75em}
\end{table*}

\subsection{Pruning Strategies Ablation: \textsc{IntraP} vs.\ \textsc{InterP}}
\label{subsec:ablation_pruning}
We ablate the two pruning strategies, Intra-Group (\textsc{IntraP}) 
and Inter-Group (\textsc{InterP}), introduced in~\autoref{subsec:murphy_with_pruning}, 
training Qwen3-1.7B on 1{,}000 KodCode samples under each strategy 
(\autoref{tab:murphy_pruning}). Both strategies reduce the number of 
retained trajectories per rollout tree (per prompt) from 72 to roughly half, lowering  the per-step optimization cost without altering the rollout procedure 
itself. The two strategies differ in which trajectories they retain, 
and this leads to different multi-turn behaviors: \textsc{IntraP} 
improves single-turn performance (Iter-1) on both benchmarks but 
regresses by \textbf{2--2.3 pp} at Iter-3, whereas \textsc{InterP} preserves 
multi-turn performance within \textbf{0.4--0.8 pp} of unpruned \method. 
\textsc{InterP} therefore offers the better cost-quality trade-off, 
retaining the multi-turn self-correction signal that pruning could 
otherwise discard.

\begin{table*}[t]
\caption{Pruned variants of \method~on Qwen3-1.7B. All variants 
generate 72 rollout trajectories per query before pruning; the 
\textit{Retained} column reports the number of trajectories kept after 
pruning per prompt, which determines how many sequence-level terms enter the 
GRPO loss. $\Delta_3$ compares \textbf{Iter-3} against unpruned 
\method~(\textsc{MaRS}) \textbf{Iter-3};~\textcolor{red}{red} 
indicates regression.}
\small
\setlength{\tabcolsep}{5pt}
\centering
\resizebox{0.95\textwidth}{!}{%
\begin{tabular}{l c ccc ccc}
\toprule
\textbf{Model \& Strategy} & \textbf{Retained}
& \multicolumn{3}{c}{\textbf{HumanEval (\%)}} 
& \multicolumn{3}{c}{\textbf{MBPP (\%)}} \\
\cmidrule(lr){3-5} \cmidrule(lr){6-8}
 &  & \textbf{Iter-1} & \textbf{Iter-3} & \textbf{$\Delta_3$}
    & \textbf{Iter-1} & \textbf{Iter-3} & \textbf{$\Delta_3$} \\
\midrule
\midrule
\method~(\textsc{MaRS}) & 72
& \underline{79.67 $\pm$ 3.01} & \textbf{86.58 $\pm$ 1.06} & \deltaz
& \underline{44.73 $\pm$ 0.50} & \textbf{62.00 $\pm$ 1.91} & \deltaz \\
\method~(\textsc{MaRS}) -- \textsc{IntraP} & 36
& \textbf{82.34 $\pm$ 1.03} & 84.28 $\pm$ 2.01 & \deltar{2.30}{40}
& \textbf{45.73 $\pm$ 0.50} & 59.87 $\pm$ 1.70 & \deltar{2.13}{38} \\
\method~(\textsc{MaRS}) -- \textsc{InterP} & 40
& 77.43 $\pm$ 2.20 & \underline{86.17 $\pm$ 0.70} & \deltar{0.41}{20}
& 44.53 $\pm$ 0.90 & \underline{61.20 $\pm$ 1.39} & \deltar{0.80}{25} \\
\bottomrule
\end{tabular}%
}
\label{tab:murphy_pruning}
\end{table*}

\textbf{Efficiency gains.} To localize the speedup, we 
profile per-step timing averaged across an end-to-end \textsc{InterP} 
run against unpruned \method~under matched configurations (Qwen3-1.7B, 
8$\times$H100 GPUs, 72 rollouts per prompt: 8 in turn-1 and up to 64 in turn-2); see~\autoref{tab:pruning_timing}. 
Generation (vLLM) and reward computation (code execution) are 
essentially unchanged ($-0.3\%$ and $-8.0\%$ respectively), because 
pruning is applied \emph{after} rollouts are collected. The savings 
concentrate entirely in the optimization phase, where pruning reduces 
the number of sequences entering gradient computation: optimization 
time drops by $\mathbf{74.1\%}$, yielding a $\mathbf{21.4\%}$ reduction in average 
per-step training time. This confirms that \textsc{InterP}'s 
computational benefit comes from avoiding gradient updates on 
uninformative trajectories, not from shortening rollouts, an important 
property, since aggressive rollout-time pruning would risk discarding 
feedback signal that later turns rely on.\looseness=-1

%% file: section/related_work.tex
\section{Related Work}
\label{sec:related_work}

\paragraph{LLM Agents for Software Development.} 
Recent studies~\citep{code-gen-llm-survey, ldb} investigate LLM agents for code generation, bug fixing, and code migration. A central factor behind their progress is inference-time iterative frameworks~\citep{reflexion, dot}, which leverage execution feedback and self-reflection to refine candidate programs~\citep{sweagent, agentless}. While such methods enhance inference pipelines, they leave the base model unchanged. In contrast, our work improves the model itself through training-time optimization, strengthening the reasoning and self-correction abilities that agentic frameworks depend on.

\textbf{RLVR for LLM Reasoning.} 
GRPO~\citep{grpo} renewed interest in RL as an efficient alternative to
PPO~\citep{ppo} for post-training LLMs on verifiable objectives, and
follow-up variants~\citep{vapo, dapo, vcppo, gspo} improve stability,
convergence, or shift optimization from token-level to sequence-level.
These methods, however, remain single-turn. A separate line of work
incorporates execution feedback during training but does so by introducing
auxiliary machinery: $\mu$\textsc{Code}~\citep{multiturncodegen} trains a
separate verifier model that scores candidate solutions and relabels
trajectories via expert iteration; RLEF~\citep{rlef} optimizes multi-turn
rollouts with a PPO critic over execution-based rewards; and
\textsc{ReVeal}~\citep{reveal} co-trains a self-verifier within the policy
and uses turn-aware optimization to drive co-evolution of generation and
verification. Concurrent work on multi-turn GRPO for tool-use
agents~\citep{mtgrpo} introduces turn-level rewards based on predefined
interaction signals (e.g., tool-execution success). In code generation,
execution can provide a scalar reward at each turn, but such intermediate
rewards are often sparse or myopic: a failed program may receive low reward
while exposing a stack trace, edge case, or failing test that enables a
later correction. Thus, \method~uses retrospective tree-based propagation
to assign credit according to downstream refinements induced by each
feedback context. Concurrent to our work, \textsc{TreeGRPO}~\citep{treegrpo}
also introduces tree-structured credit assignment for GRPO-style
post-training, but in a different setting: visual generative models, where
the tree branches within a single denoising trajectory and leaf advantages
are propagated to denoising edges. In contrast, \method~constructs
feedback-conditioned trees across multi-turn code-refinement attempts,
propagates execution rewards before local group-relative normalization, and
studies both optimistic (\textsc{MaRS}) and expectation-like
(\textsc{MeRS}) backups. We provide a detailed comparison in
App~\ref{app:tree-credit}. Across these lines of work, \method~differs by
extending GRPO to multi-turn code generation through feedback-conditioned
rollout trees and retrospective reward propagation, without a learned
verifier, an auxiliary critic, or predefined dense turn-level rewards.
Empirically, this minimal extension matches or outperforms
$\mu$\textsc{Code} and \textsc{ReVeal} across our benchmarks
(\autoref{sec:experiments}). See~\autoref{app:extended_related_work} for
an extended discussion.
\looseness=-1

%% file: section/conclusion.tex
\section{Conclusion \& Limitations}
We introduced \method, a multi-turn extension of GRPO that addresses 
the credit-assignment mismatch in feedback-conditioned code generation. 
\method~constructs feedback-conditioned rollout trees, then propagates 
rewards backward so that earlier failed attempts receive credit for the 
informative feedback they surface. Across three benchmarks and two 
model families, \method~delivers up to 6\% absolute pass@1 gains over 
the strongest prior multi-turn execution-feedback methods, with the 
largest improvements on harder problems where iterative self-correction 
matters most. These results suggest that the standard GRPO recipe can 
be made feedback-aware through a structural change to credit 
assignment, without learned verifiers or auxiliary critics. \textbf{Limitations:}
\method's multi-turn rollouts increase training cost relative to GRPO.
\textsc{InterP} mitigates this cost, reducing average per-step training
time, but multi-turn training
remains more expensive due to rollout-tree expansion. In our experiments,
we use vLLM for rollout generation; shared prefixes across parent and child
nodes could further enable prefix caching and KV-cache reuse, which we leave
to future work. Our evaluation focuses on code generation, where execution
provides clean feedback. Extending \method~to noisier feedback signals,
longer refinement horizons, broader agentic tasks, and adaptive turn or
rollout budgets is a natural next step.\looseness=-1

%% file: section/appendix.tex
\appendix

\begin{center}
\rule{\textwidth}{1pt}\\[0.75em]
{\LARGE\bfseries Appendix}\\[0.5em]
\rule{\textwidth}{1pt}
\end{center}

\begingroup
\setcounter{tocdepth}{2}
\renewcommand{\contentsname}{}
\tableofcontents
\endgroup

\newpage


    
    
    
    
    
    


    

\section{Extended Related Work}
\label{app:extended_related_work}
\paragraph{LLM Agents for Software Development.} Recent works~\citep{code-gen-llm-survey, ldb} have explored LLM agents for programming tasks such as code generation, bug fixing, and code migration. A key driver of progress in these domains has been inference-time iterative frameworks~\citep{reflexion, dot}, which leverage execution feedback to generate self-reflections for refining candidate programs~\citep{sweagent, agentless}. While these approaches underscore the value of iterative feedback and scaffolding, they primarily enhance the inference pipeline rather than the underlying model. Our work takes a complementary direction: we improve the reasoning and self-correction abilities of LLMs themselves through training-time optimization, thereby strengthening the base models that agentic frameworks depend on.\looseness=-1

\paragraph{Reinforcement Learning with Verifiable Rewards for LLM Reasoning.} Reinforcement learning (RL) is a popular paradigm for post-training LLMs to improve reasoning and align outputs with verifiable objectives. GRPO~\citep{grpo} revived interest in RL as an efficient alternative to PPO~\citep{ppo}, offering comparable reasoning performance with far lower computational cost. Subsequent variants~\citep{vapo, dapo, vcppo, gspo} focus on stabilizing training, improving convergence, or shifting optimization from token-level to sequence-level. However, these algorithms remain tailored to single-turn tasks, optimizing models to produce one-shot completions without iterative refinement. Recent RLVR methods extend RL to multi-turn agents across domains such as search~\citep{reSearch, searchR1}, tool use~\citep{bespoke_improving_multi_turn_tool_use, toolN1}, and code generation~\citep{multiturncodegen, rlef}. These methods typically compute advantage by summing outcome and turn-level rewards, which limits temporal credit propagation. In contrast, \method~delays credit assignment until a trajectory is complete, propagating rewards backward from successful states using a structured credit assignment criterion that preserves temporal consistency. Our work is most closely related to $\mu$Code~\citep{multiturncodegen} and RLEF~\citep{rlef}, which also train LLMs with execution feedback. $\mu$Code jointly trains a generator and a learned verifier that scores multi-turn code solutions, while RLEF refines generations via PPO grounded in execution results. However, both approaches require auxiliary value functions or verifier LLMs, significantly increasing computational overhead and data acquisition cost. $\mu$Code further depends on its verifier at inference for Best-of-N selection, introducing additional latency. These design choices make direct comparison impractical and obscure the effect of the RL formulation itself. In contrast, \textsc{Murphy} achieves comparable grounding in execution feedback and iterative refinement by extending GRPO to the multi-turn setting, while preserving its simplicity, efficiency, and architectural minimalism.\looseness=-1

\subsection{Relation to Tree-Structured GRPO and Backup Theory}
\label{app:tree-credit}

\textsc{TreeGRPO}~\citep{treegrpo} is closely related in spirit to
\method: both methods replace trajectory-level credit with tree-structured
credit assignment before applying a GRPO-style update. However, the two methods
differ in the source of branching, the object being propagated, and the point at
which normalization is applied. \textsc{TreeGRPO} targets diffusion/flow
post-training, where branching occurs within a single denoising trajectory. It
first computes normalized leaf advantages and then propagates them backward to
obtain per-edge advantages using a log-probability weighted average over child
branches. \method~instead targets multi-turn code generation, where branching is
induced by executor feedback: failed programs create feedback-conditioned prompts
whose descendants are later refinements. We therefore propagate execution rewards
through the tree and then compute local group-relative advantages within each
feedback-conditioned response group.

Tree-structured GRPO methods differ primarily in what quantity is backed up
and how descendants are aggregated. \textsc{TreeGRPO}~\citep{treegrpo}
propagates leaf advantages backward through a denoising tree using a
policy-weighted average:
\[
A(v) = \sum_{u \in C(v)} w_v(u) A(u),
\qquad
w_v(u) =
\frac{\pi_{\theta_{\mathrm{old}}}(u \mid v)}
{\sum_{u' \in C(v)} \pi_{\theta_{\mathrm{old}}}(u' \mid v)}.
\]
In contrast, \method~propagates execution rewards through a
feedback-conditioned refinement tree and computes local GRPO advantages only
after this reward propagation step. This leads to two complementary backups:
\[
\textsc{MeRS}: \qquad
\tilde r(v)
=
\frac{
r(v) + \gamma \cdot \frac{1}{|C(v)|}\sum_{u \in C(v)} \tilde r(u)
}{
Z(v)
},
\]
and
\[
\textsc{MaRS}: \qquad
\tilde r(v)
=
\max\left(r(v), \max_{u \in C(v)} \tilde r(u)\right).
\]

Here \(Z(v)\) denotes the local normalization factor used in the reward backup.
Unlike the policy-weighted averaging rule in \textsc{TreeGRPO}, this
normalization factor is tied to the realized refinement subtree below \(v\).
It may therefore depend on the number of sampled descendants, the rollout depth,
and the feedback-induced branching pattern. Consequently, \textsc{MeRS} should
not be interpreted as the same estimator as the \textsc{TreeGRPO} advantage
backup. Rather, it is a sample-dependent reward propagation rule that smooths
execution rewards over the refinements actually generated under a given feedback
context.

\textsc{MeRS} is closest in spirit to the aggregation rule in
\textsc{TreeGRPO}: both combine information from child branches before applying
a GRPO-style update. The analogy, however, is only structural. In
\textsc{TreeGRPO}, the backed-up quantity is a normalized advantage and the
weights are induced by the old policy over denoising branches. In \method, the
backed-up quantity is an execution reward, and the normalization term \(Z(v)\)
is part of the sampled reward-propagation procedure.

Averaging over child branches can be viewed as a variance-reducing operation
conditional on a fixed realized child set. Specifically, if \(C(v)\) and weights
\(w_v\) are fixed, and if child branch values \(X_u\) are conditionally
independent with common variance \(\sigma^2\), then
\[
\mathrm{Var}\!\left[
\sum_{u \in C(v)} w_v(u) X_u
\,\middle|\,
C(v), w_v
\right]
=
\sigma^2 \sum_{u \in C(v)} w_v(u)^2.
\]
For uniform weights this becomes \(\sigma^2/|C(v)|\), conditional on the
realized child set. Since the refinement tree, the number of descendants, and
the normalization term \(Z(v)\) are themselves determined by the sampled
executor-feedback process, this conditional calculation should not be read as an
unconditional variance or unbiasedness guarantee. Instead, it provides intuition
for why \textsc{MeRS} behaves as a smoothing backup over the realized
refinement tree, whereas \textsc{MaRS} behaves as an optimistic backup that
propagates the best observed execution reward.

\textsc{MaRS}, by contrast, is an optimistic best-descendant backup. For
binary correctness rewards,
\[
\mathbb{E}\!\left[\max_{u \in C(v)} R(u)\right]
=
\Pr\!\left(\exists u \in C(v): R(u)=1\right),
\]
and, under conditional independence, this equals
\[
1 - \prod_{u \in C(v)} (1-p_u)
\]
This objective is well aligned with sparse-reward code self-correction: a
failed attempt should receive credit if the feedback it induces enables at
least one later refinement to solve the task. Prior analysis of Monte Carlo
tree-search backups similarly shows that max-style backups can preserve rare
high-value trajectories that averaging may dilute~\citep{khandelwal2016complex}.

This distinction also clarifies the roles of \textsc{MeRS} and \textsc{MaRS}.
\textsc{MeRS} is closest to \textsc{TreeGRPO}'s aggregation rule: both estimate
the value of an internal node by averaging over sampled descendants, and both
can be interpreted as reducing variance relative to relying on a single sampled
continuation. \textsc{MaRS}, by contrast, is an optimistic best-descendant
backup. This is motivated by the sparse and existential nature of code
self-correction: a failed attempt can be useful if it exposes feedback that
enables at least one later refinement to solve the task. Prior work on backup
strategies in Monte Carlo tree search similarly observes that averaging can
drown out rare high-value trajectories, while max-style backups can be
beneficial when successful trajectories are scarce~\citep{khandelwal2016complex}.
Thus, \textsc{MeRS} provides a lower-variance expectation-like backup, whereas
\textsc{MaRS} is better aligned with best-of-$k$ and pass@$k$ style success
criteria in sparse-reward code generation.

Recent work on \textsc{MaxRL}~\citep{maxrl} provides
a complementary perspective: in correctness-based tasks, likelihood-oriented
objectives emphasize successful rollouts and higher-order pass@$k$ events rather
than only expected pass@1 reward. While MaxRL is developed for single-turn
binary-outcome settings, it supports the intuition behind \textsc{MaRS}: when
success is rare, training signals that preserve and amplify successful
descendants can be preferable to averaging them away. Our contribution is to
instantiate this principle in a feedback-conditioned multi-turn GRPO setting,
where earlier failed attempts are credited according to the downstream
refinements they make possible.

\section{GRPO: Objective and Additional Details}
\label{app:grpo_additional_details}

\paragraph{Notation.} We denote the model policy by $\pi_{\theta}(\cdot \mid \cdot)$ and the reference (older) policy by $\pi_{\theta_{\text{old}}}(\cdot \mid \cdot)$. Let $G$ be the number of generations per prompt, $\mathcal{P}(Q)$ the distribution over input prompts/questions $Q$, and $O$ the output space. For a given prompt $q \sim \mathcal{P}(Q)$, the reference policy produces a set of $G$ responses, forming a response group $\{o_{q,1}, o_{q,2}, \ldots, o_{q,G}\}$. Each generation $o_{q,i} \in O$ corresponds to a full output trajectory, where $o_{q,i,t}$ denotes the $t$-th token and $o_{q,i,<t}$ the prefix up to (but excluding) token $t$. We write $|o_{q,i}|$ for the sequence length of the $i$-th generation. For a given prompt $q$, the reward model assigns a scalar score to each response in the group, yielding $\mathbf{r}_q = \{r_{q,1}, r_{q,2}, \ldots, r_{q,G}\}$. Moreover, for prompt $q$, the advantage associated with the $t$-th token of the $i$-th generation is defined as $\hat{A}_{q, i,t} = (r_{q, i} - \mu(\mathbf{r}_q)) / \sigma(\mathbf{r}_q)$, where $\mu(\mathbf{r}_q)$ and $\sigma(\mathbf{r}_q)$ denote the mean and standard deviation of the group rewards, respectively. Finally, $D_{\mathrm{KL}}(\pi_\theta|| \pi_{\theta_{\text{ref}}})$ denotes the KL divergence between the current and reference policies, computed over all tokens in the generated sequences. The GRPO training objective is presented in \autoref{def:grpo_objective}. 

\begin{definitionbox}[label=def:grpo_objective]{(GRPO Objective) }
\begin{align*}
\mathcal{J}(\theta) =\ & \mathbb{E}_{q \sim \mathcal{P}(Q),\ \{o_{(q, i)}\}_{i=1}^G \sim \pi_{\theta_{\text{old}}}(O|q)} \Big[\frac{1}{G} \sum_{i=1}^G  \frac{1}{|o_{q, i}|}\sum_{t=1}^{|o_{q,i}|} \min \Big( R_{\theta}(q, i, t)\hat{A}_{q, i,t}, \\
&\text{clip} \Big( R_{\theta}(q, i, t),\,
1 - \epsilon,\,
1 + \epsilon
\Big) \hat{A}_{q, i,t} \Big) \Big] - \beta D_{\text{KL}}(\pi_\theta \,\|\, \pi_{\text{ref}})
\end{align*}

\vspace{-5pt}
Where,
\vspace{-8pt}
\begin{align*}
    R_{\theta}(q, i, t) &= \frac{\pi_\theta(o_{q,i,t} \mid q, o_{q, i,<t})}{\pi_{\theta_{\text{old}}}(o_{q, i,t} \mid q, o_{q, i,<t})} \\
    \hat{A}_{q, i,t} &= \frac{(r_{q, i} - \mu(\mathbf{r}_q))}{ \sigma(\mathbf{r}_q)}
\end{align*}
\end{definitionbox}

\section{Multi-turn Rollout Formalism (Detailed)}
\label{app:murphy_formalism}

This appendix section provides a detailed treatment of the rollout-tree construction and indexing used throughout the main paper.

We define a feedback-conditioned rollout tree that captures how model generations evolve across multiple turns of interaction with the environment. Let $s$ denote the turn index, $S$ the total number of turns, and $G_s$ the number of generations per prompt at turn $s$. 

\colorbox{ngreen}{\emph{Turn 1}:} In the first turn ($s=1$), the model receives an input prompt $q_{(1)}$ sampled from $\mathcal{P}(Q)$ and generates $G_1$ candidate programs which forms a response group $\{o_{\{q_{(1)}, 1\}}, o_{\{q_{(1)},2\}}, \ldots, o_{\{q_{(1)},G_1\}}\}$. Note that $\{ o_{\{q_{(1)}, j\}} \}_{j=1}^{G_1} \sim \pi_{\theta_{\text{old}}}(\cdot \mid q_{(1)})$. Each generation is executed against its associated test suite, producing a numerical reward $r_{\{q_{(1)}, j\}}$ and environment feedback $f_{\{q_{(1)}, j\}}$.  The reward represents the proportion of test cases passed. The environment feedback contains the specific unit tests that passed or failed, along with any corresponding error messages. These $G_1$ generations form the first layer of output nodes in the rollout tree.\looseness=-1 

\colorbox{ngreen}{\emph{Turn 2}:} For each generation $j$ in turn 1 that fails to achieve the maximum reward (which equals 1 in our setting, since it represents the proportion of test cases passed), the corresponding feedback is appended to the original prompt and the prior output from turn $1$ to form a feedback-conditioned prompt: $q_{(2, j)} = [\ q_{(1)},\ o_{\{q_{(1)}, j\}},\ f_{\{q_{(1)}, j\}}\ ]$ where $[\,\cdot\,]$ denotes textual concatenation. The model is then re-invoked to generate $G_2$ new candidate solutions:  $\{o_{\{q_{(2, j)},k\}}\}_{k=1}^{G_2} \sim \pi_{\theta_{\text{old}}}(\cdot \mid q_{(2, j)})$. Each of these generations is evaluated to obtain $(r_{\{q_{(2, j)}, k\}}, f_{\{q_{(2, j)}, k\}})$, which denote the reward and feedback. These output generations represent refinements of their corresponding parent outputs $o_{\{q_{(1)}, j\}}$ and collectively form the second layer of the rollout tree. 

\colorbox{ngreen}{{\emph{Turn $s$}}:} Building on the previous turn, this procedure extends recursively to any turn $s \in \{1, \dots, S-1\}$. We define $i_{[1:s]} = i_1, \dots, i_s$ as a sequence of branch indices that trace a specific path through the tree, where $i_j$ indicates the $i$'th candidate that was selected at turn $j$. Similarly, $i_{[1:1]}$ denotes $i_1$. For each generation $o_{{q_{(s, i_{[1:s-1]})}, i_s}}$ that fails to achieve the maximum reward, we construct a feedback-conditioned prompt:
\begin{align*}
q_{(s+1, i_{[1:s]})} = [q_{(s, i_{[1:s-1]})},\, o_{{q_{(s, i_{[1:s-1]})}, i_s}},\, f_{{q_{(s, i_{[1:s-1]})}, i_s}}]
\end{align*}
where $[\cdot]$ denotes textual concatenation. The model is then re-invoked to generate $G_{s+1}$ new candidate solutions:
\begin{align*}
\{o_{\{q_{(s+1, i_{[1:s]})}, k\}} \}_{k=1}^{G_{s+1}} \sim \pi_{\theta_{\text{old}}}(\cdot \mid q_{(s+1, i_{[1:s]})})
\end{align*}
Each candidate is evaluated to obtain its reward and feedback $(r_{\{q_{(s+1, i_{[1:s]})}, k\}} ,\, f_{\{q_{(s+1, i_{[1:s]})}, k\}})$. The resulting generations, $\{o_{\{q_{(s+1, i_{[1:s]})}, k\}} \}_{k=1}^{G_{s+1}}$, form the child nodes of parent node $o_{\{q_{(s, i_{[1:s-1]})}, i_s\}}$.

A complete path from the root (the initial prompt at turn 1) to a leaf at turn $S$ (final output) can be expressed as: \looseness=-1
\begin{align*}
q_{(1)} \rightarrow o_{{q_{(1)}, i_1}} \rightarrow q_{(2, i_{[1:1]})} \rightarrow o_{{q_{(2, i_{[1:1]})}, i_2}} \rightarrow \cdots \rightarrow o_{\{q_{(S, i_{[1:S-1]})}, i_S\}}
\end{align*}
where the indices $i_1, \dots, i_S$ specify which branch is taken at each turn. Leaf nodes at turn $S$ represent the final generations obtained after completing all refinement steps. Once rewards for all turns are computed, the rewards from these terminal nodes are propagated backward through their ancestors according to the credit assignment strategies described below.

\colorbox{nblue}{\textbf{Credit assignment formalism.}} After all outputs and corresponding rewards are generated and the rollout tree is constructed, we focus on assigning credit from later turns back to earlier ones. To achieve this, we explore two distinct strategies.\looseness=-1

\colorbox{ngreen}{\emph{Max Reward Strategy (\textsc{MaRS})}:} The Max Reward Strategy is defined recursively, proceeding from the final turn back to the root. Since the final turn $S$ has no children, the rewards at this turn remain unchanged. We then consider turn $s = S - 1$. Let $o_{\{q_{(s, i_{[1:s-1]})}, i_{s}\}}$ denote a generation at turn $s = S - 1$ with an associated reward $r_{\{q_{(s, i_{[1:s-1]})}, i_{s}\}}$. If this generation already achieves the maximum reward, it has no children; otherwise, its children are defined as:\looseness=-1
\begin{align*}
    C\big(o_{\{q_{(s, i_{[1:s-1]})}, i_s\}}\big) = \{o_{\{q_{(s+1, i_{[1:s]})}, 1\}},\ \ldots,\  o_{\{q_{(s+1, i_{[1:s]})}, G_S\}}\}
\end{align*}
The corresponding set of rewards are defined as
\begin{align*}
    C_r\big(o_{\{q_{(s, i_{[1:s-1]})}, i_s\}}\big) = \{r_{\{q_{(s+1, i_{[1:s]})}, 1\}},\ \ldots,\ r_{\{q_{(s+1, i_{[1:s]})}, G_S\}}\}
\end{align*}
which defaults to zero if there are no children.
We then update the reward as: 
\begin{align*}
    r_{\{q_{(s, i_{[1:s-1]})}, i_{s}\}} &= \text{max}\Bigg(r_{\{q_{(s, i_{[1:s-1]})}, i_{s}\}},\ \text{max}\Big(C_r(o_{\{q_{(s, i_{[1:s-1]})}, i_{s}})\Big) \Bigg)
\end{align*}

This strategy assigns each node the maximum of its own reward and the best reward among its descendants. Intuitively, the descendant maximum represents the best outcome achievable through refinement, while taking the outer maximum ensures a node's credit never decreases; even when feedback fails to improve performance. This formulation captures the maximum progress achievable from any refinement path starting at that node. The procedure operates recursively in a backward pass: rewards are first updated for all nodes at turn $S-1$ based on their children's values at turn $S$. This process continues backward through the tree, with each turn $s$ receiving updated rewards based on the values from turn $s+1$, until reaching the root.

\colorbox{ngreen}{\emph{Mean Reward Strategy (\textsc{MeRS})}:} The Mean Reward 
Strategy follows the same recursive credit assignment structure as the Max Reward 
Strategy (\textsc{MaRS}), but differs in how rewards are propagated. Inspired by the 
return computation in REINFORCE~\cite{reinforce}, \textsc{MeRS} updates each node's reward by incorporating 
the discounted \emph{mean} of its children's rewards.

For a generation $o_{\{q_{(s, i_{[1:s-1]})}, i_s\}}$ at turn $s$, let 
$\mathbb{I}_{\text{unsolved}}$ be an indicator that equals 1 if the problem 
remains unsolved at this node, and 0 otherwise. The update rule is:
\begin{align*}
  r_{\{q_{(s, i_{[1:s-1]})}, i_s\}} =
  \frac{r_{\{q_{(s, i_{[1:s-1]})}, i_s\}} + \gamma \cdot \overline{C}_r\Big(o_{\{q_{(s, i_{[1:s-1]})}, i_s\}}\Big)}
  {\mathbb{I}_{\text{unsolved}} \cdot (S - s) + 1}
\end{align*}
where $\gamma \in [0, 1]$ is a discount factor controlling the influence of 
descendant rewards, and $\overline{C}_r(\cdot)$ denotes the mean reward over 
children of unsolved nodes (children of solved nodes do not exist since 
they represent terminal states).

The denominator equals $S - s + 1$ for unsolved nodes and $1$ for solved nodes. 
This depth-based normalization serves two purposes: (1) it prevents rewards 
from growing unboundedly during backward propagation, as unsolved nodes can 
potentially accumulate discounted contributions from up to $S - s$ future turns; 
and (2) it ensures fair comparison between nodes that solve at different turns, a 
node that solves immediately at turn $s$ retains its full reward, while a node 
that fails but has successful descendants receives appropriately scaled credit 
that accounts for the additional refinement steps required. All other aspects 
of the recursive procedure remain identical to \textsc{MaRS}.

\colorbox{ngreen}{\emph{\textsc{MaRS} vs \textsc{MeRS}}:} \textsc{MaRS} propagates the maximum descendant reward to the earlier turns, emphasizing peak performance, whereas \textsc{MeRS} propagates the mean reward, emphasizing stability and overall consistency. \textsc{MaRS} captures best-case improvement, while \textsc{MeRS} provides a smoother estimate of expected progress. Together, they offer complementary views of feedback-driven credit assignment.\looseness=-1

\colorbox{nblue}{\emph{\method~Objective}:} Once the rewards are reassigned according to the chosen credit assignment strategy (\textsc{MaRS} or \textsc{MeRS}), advantages are computed similar to standard GRPO. The adjusted rewards serve as the basis for computing normalized advantages at each turn. Conditioned on a prompt $\tilde{q}$ and for $G_s$ generations, we normalize each reward by subtracting the mean reward and dividing by the standard deviation of the rewards obtained across the $G_s$ generations, yielding the normalized advantage $\hat{A}^{\method}_{\tilde{q}, i, t}$. Additionally, as defined earlier, each $i$-th generation at turn $s$, $o_{\{q_{(s, i_{[1:s-1]})}, i\}} \in O$, corresponds to a complete output trajectory, where $o_{\{q_{(s, i_{[1:s-1]})}, i, t\}}$ denotes the $t$-th token and $o_{\{q_{(s, i_{[1:s-1]})}, i, <t\}}$ the prefix up to (but excluding) token $t$. We denote the sequence length of the $i$-th generation by 
$|o_{\{q_{(s, i_{[1:s-1]})}, i\}}|$. At each turn, the GRPO objective is applied using these credit-adjusted advantages. This per-turn optimization allows \method~to incorporate feedback from later turns into earlier updates, effectively extending GRPO to a multi-turn setting. Finally, the divergence between the current and reference policies is captured by $D_{\mathrm{KL}}(\pi_\theta \,\|\, \pi_{{\text{ref}}})$, computed over all tokens in the generated sequences. The resulting optimization objective, which integrates credit-assigned rewards, normalized advantages, and KL regularization at each turn, defines the \method~objective, distinguishing it from the standard GRPO formulation. The full \method~objective is presented in \autoref{def:method_objective}. \looseness=-1

\begin{definitionbox}[label=def:method_objective]{(\method ~Objective) }
\begin{align*}
    \mathcal{J}_{\method}(\theta) &= \mathbb{E}_{q\sim \mathcal{P}(Q)}[\sum_{s, i_1, \ldots i_S}\mathcal{J}_{q_{(s, i_{[1:s]})}}(\theta)]
\end{align*}
Where the per prompt objective at each turn $s$ is:
\vspace{-4pt}
\begin{align*}
\mathcal{J}_{q_{(s, i_{[1:s-1]})}}(\theta) &=\mathbb{E}_{\{o_{(\tilde{q}, i)}\}_{i=1}^{G_s} \sim \pi_{\theta_{\text{old}}}(O|\tilde{q})}\Bigg[
\sum_{i=1}^{G_s} \frac{1}{G_s |o_{\{\tilde{q} ,i\}}|} \sum_{t=1}^{|o_{\{\tilde{q}, i\}}|}\Bigg(\min\Big(R_\theta(\tilde{q},i,t) \hat{A}^{\method}_{\tilde{q}, i,t}, \\
&\quad\quad\quad\quad\text{clip}\left(
R_\theta(\tilde{q},i,t),\,
1 - \epsilon,\,
1 + \epsilon \right) \cdot \hat{A}^{\method}_{\tilde{q}, i,t}\Big) - \beta D_{\mathrm{KL}}(\pi_\theta \,\|\, \pi_{{\text{ref}}})\Bigg)
\Bigg] 
\end{align*}
\begin{align*}
   \text{with,~} \tilde{q} &= q_{(s, i_{[1:s-1]})} 
\end{align*}
\begin{align*}
R_\theta(\tilde{q},i,t)&=\frac{\pi_\theta(o_{\{\tilde{q}, i, t\}} \mid \tilde{q}, o_{\{\tilde{q}, i, <t\}})}{\pi_{\theta_{\text{old}}}(o_{\{\tilde{q}, i, t\}} \mid \tilde{q}, o_{\{\tilde{q}, i, <t\}})}
\end{align*}
And,
$\hat{A}^{\method}_{\tilde{q}, i, t}$ denotes the advantage. 
\end{definitionbox}

\paragraph{Note.} The design of \method~is broadly applicable across a range of RLVR algorithms, including PPO~\citep{ppo} and various extensions of GRPO~\citep{rloo, dapo, vapo}. In this work, we focus on GRPO due to its strong empirical performance in aligning LLMs~\citep{dsr1, Qwen3}. Extending \method~to other RLVR variants is conceptually straightforward, as it builds on the same underlying principles. While our experiments primarily focus on code generation, where rich, verifiable feedback is readily available, the framework can naturally extend to other domains such as mathematics or logical reasoning, provided suitable forms of feedback are accessible.

\section{Sensitivity to Multi-Iteration Scaffolds}
\label{app:scaffold_sensitivity}

\method~is explicitly designed to train models to incorporate environment feedback and self-correct over multiple turns. In multi-iteration evaluation—typical of agentic settings where feedback is present—\method~yields substantially larger gains. This behavior is expected and reflects the objective of \method, which is to improve a model’s ability to utilize feedback across turns rather than to optimize single-pass generation.

To assess sensitivity to the choice of multi-iteration scaffold, we conduct additional experiments using the Qwen3-1.7B model. We compare the base model, the GRPO-trained model, and the \method-trained model under three distinct scaffolds: Reflexion~\citep{reflexion}, LATS~\citep{lats}, and DoT~\citep{dot}. Importantly, no additional training is performed. We evaluate the same GRPO-MT checkpoint reported in ~\autoref{fig:humaneval_mbpp} and the same \method~(\textsc{MaRS}, \textsc{IntraP}) checkpoint reported in Table~\ref{tab:murphy_pruning}. All scaffolds use their standard configurations with three iterations. We report mean accuracy and standard deviation over three independent runs on HumanEval.

\begin{table*}[h!]
\caption{Sensitivity of Qwen3-1.7B performance on HumanEval to different multi-iteration scaffolds. No additional training is performed.}
\centering
\small
\setlength{\tabcolsep}{8pt}
\begin{tabular}{lccc}
\toprule
\textbf{Model} & \textbf{Reflexion} & \textbf{LATS} & \textbf{DoT} \\
\midrule
Qwen3-1.7B (Base) & 80.07 $\pm$ 1.27 & 78.95 $\pm$ 1.29 & 78.35 $\pm$ 3.88 \\
Qwen3-1.7B GRPO-MT & 81.14 $\pm$ 1.62 & 81.34 $\pm$ 2.51 & 82.92 $\pm$ 0.47 \\
Qwen3-1.7B \method~(\textsc{MaRS}, \textsc{IntraP}) & \textbf{84.28 $\pm$ 2.01} & \textbf{86.89 $\pm$ 0.42} & \textbf{85.66 $\pm$ 0.43} \\
\bottomrule
\end{tabular}
\label{tab:scaffold_sensitivity}
\end{table*}

Across all scaffolds, the \method-trained model consistently outperforms both the GRPO-MT trained model (trained under a matched compute budget) and the base model. These results indicate that the gains from \method~are not dependent on any particular inference-time scaffold, but instead reflect improved feedback utilization learned during training.

\section{Reflexion}
\label{app:reflexion}
Reflexion~\citep{reflexion} is an inference-time iterative framework designed to improve reasoning through repeated interaction with feedback from an external environment. It employs three agents: an actor ($M_a$), an evaluator ($M_e$), and a self-reflection module ($M_{sr}$), which operate cyclically until a termination condition is met. For code generation, the process proceeds as follows:

\begin{enumerate}
    \item \textbf{Actor step:} The actor $M_a$ receives an input and generates an output (e.g., a code snippet).
    \item \textbf{Evaluation step:} The evaluator $M_e$ scores the output (e.g., the percentage of  unit tests passed).
    \item \textbf{Self-reflection step:} If the score is insufficient, the self-reflection module $M_{sr}$ diagnoses the issue, proposes a fix, and appends both the failed output and the suggested correction to the input context. The updated input is then fed back to the actor, and the cycle repeats until either the task succeeds or a maximum number of iterations is reached.
\end{enumerate}

In most implementations, the actor and the self-reflection module are instantiated by the same underlying language model. The self-reflection stage thus corresponds to the model reasoning over its own prior outputs augmented with feedback from the evaluator (or executor) and the previous input, output pairs, to generate improved responses in subsequent iterations.

\section{Implementation Details}
\label{app:implementation_details}

We implement our framework on top of TRL \citep{vonwerra2022trl}, which provides efficient distributed training and a modular implementation of GRPO. We integrate TRL with vLLM for fast inference and large-scale rollout execution, enabling scalable multi-turn training in our experiments. Prompts used to train \method~are listed in \autoref{app:prompt_examples}. All experiments use publicly available datasets. The base models (Qwen3, OLMo) are available for research use under the Apache 2.0 license.

\subsection{Model Size and Compute Budget}
All experiments were conducted on $8 \times$ NVIDIA H100 GPUs. Our implementation builds on HuggingFace's TRL\footnote{\url{https://huggingface.co/docs/trl/en/index}}. For efficiency, 2 GPUs were allocated for inference via vLLM, while the remaining 6 GPUs handled model updates. Checkpoints were saved every 50 steps, and for all baselines, we selected the checkpoint corresponding to one epoch.

\subsection{Hyperparameters}
We set the KL regularization coefficient to $\beta = 0.04$, the learning rate to $10^{-6}$, and weight decay to $0.1$ for both GRPO-MT and \method~variants. Unless otherwise specified, \method~uses two turns. In GRPO-MT, we sample 72 rollouts per prompt in turn~1; failed rollouts receive execution feedback and are extended in turn~2. For \method, we sample 8 rollouts per prompt in turn~1 and up to 8 per prompt in turn~2, yielding at most 64 rollouts in the second turn. For $\mu$\textsc{Code} and \textsc{ReVeal} we use the author recommended hyper-parameters for training. 

Following~\cite{Qwen3}, we use a temperature of 0.6 and top-$p$ of 0.95 for all inference experiments.

\subsection{Package Parameters}
We use the Reflexion~\citep{reflexion} framework to evaluate all trained models. The number of iterations is swept over \(\{1,3\}\), and \texttt{max-tokens} is set to each model's maximum generation length. Models are hosted via vLLM. Since all models fit on a single H100 GPU, we set \texttt{data-parallel-size} to 8 and enable prefix caching to accelerate evaluation. We use the following commands to install the appropriate packages: 

\begin{verbatim}
pip install uv && \
uv pip install trl==0.19.1 && \
uv pip install gunicorn==20.1.0 && \ 
uv pip install fastapi==0.115.12 
uv pip install uvicorn==0.34.2 && \
uv pip install aiohttp==3.11.18 
uv pip install astunparse==1.6.3 
uv pip install jsonlines tenacity && \
uv pip install vllm==0.8.5.post1
\end{verbatim}

\subsection{Additional Experiments}

\begin{table*}[!htbp]
\caption{Performance of OLMo-2-1124-7B-Instruct, Qwen3-1.7B and Qwen3-4B variants on HumanEval and MBPP benchmarks. GRPO-MT corresponds to the multi-turn baseline. \method\ (Ours) is highlighted. $\Delta_3$ denotes the difference between the \textbf{Iter-3} performance of the GRPO-MT/$\mu$\textsc{Code}/\method-trained models and that of \textbf{Base (Iter-3)}, within each model block; \textcolor{green}{green} indicates improvement (darker = larger gain), \textcolor{red}{red} indicates regression.}
\small
\setlength{\tabcolsep}{4.5pt}
\centering
\resizebox{\textwidth}{!}{%
\begin{tabular}{l ccc ccc}
\toprule
\textbf{Model}
& \multicolumn{3}{c}{\textbf{HumanEval (\%)}} 
& \multicolumn{3}{c}{\textbf{MBPP (\%)}} \\
\cmidrule(lr){2-4} \cmidrule(lr){5-7}
 & \textbf{Iter-1} & \textbf{Iter-3} & \textbf{$\Delta_3$}
    & \textbf{Iter-1} & \textbf{Iter-3} & \textbf{$\Delta_3$} \\

\midrule
\midrule
\textbf{OLMo-2-1124-7B-Instruct} \\
\quad Base   
& 37.20 $\pm$ 0.86 & \underline{46.04 $\pm$ 0.43} & \deltaz
& 19.90 $\pm$ 0.42 & 29.60 $\pm$ 0.28 & \deltaz \\

\quad GRPO-MT
& 39.02 $\pm$ 0.00 & 41.87 $\pm$ 0.93 & \deltar{4.17}{40}
& 28.33 $\pm$ 0.31 & 33.60 $\pm$ 0.69 & \deltag{4.00}{44} \\

\quad $\mu$Code~\citep{multiturncodegen}   
& 35.17 $\pm$ 0.35 & 41.46 $\pm$ 1.06 & \deltar{4.58}{43}
& 22.47 $\pm$ 0.42 & 31.60 $\pm$ 0.60 & \deltag{2.00}{34} \\

\oursrow
\quad \method\ - \textsc{MaRS}
& \textbf{45.53 $\pm$ 0.70} & \textbf{52.24 $\pm$ 1.96} & \deltag{6.20}{57}
& \underline{29.33 $\pm$ 0.90} & \textbf{39.67 $\pm$ 1.29} & \deltag{10.07}{80} \\

\midrule
\midrule
\textbf{Qwen3-1.7B} \\
\quad Base   
& \underline{74.19 $\pm$ 1.95} & 80.07 $\pm$ 1.27 & \deltaz
& 43.47 $\pm$ 0.64 & 53.93 $\pm$ 3.52 & \deltaz \\

\quad GRPO-MT   
& 70.33 $\pm$ 0.93 & 82.11 $\pm$ 0.35 & \deltag{2.04}{32}
& \textbf{46.00 $\pm$ 0.40} & \underline{58.53 $\pm$ 1.03} & \deltag{4.60}{44} \\

\quad $\mu$Code~\citep{multiturncodegen}   
& 73.58 $\pm$ 1.27 & \underline{83.34 $\pm$ 0.70} & \deltag{3.27}{39}
& 44.00 $\pm$ 0.69 & 56.20 $\pm$ 0.60 & \deltag{2.27}{34} \\

\oursrow
\quad \method\ - \textsc{MaRS}
& \textbf{79.67 $\pm$ 3.01} & \textbf{86.58 $\pm$ 1.06} & \deltag{6.51}{59}
& \underline{44.73 $\pm$ 0.50} & \textbf{62.00 $\pm$ 1.91} & \deltag{8.07}{68} \\

\midrule
\midrule
\textbf{Qwen3-4B} \\
\quad Base   
& \underline{90.04 $\pm$ 3.13} & 93.49 $\pm$ 0.93 & \deltaz
& \underline{52.13 $\pm$ 0.42} & 70.93 $\pm$ 1.01 & \deltaz \\

\quad GRPO-MT
& 89.00 $\pm$ 1.50 & 93.80 $\pm$ 1.00 & \deltag{0.31}{22}
& 51.50 $\pm$ 0.40 & 71.40 $\pm$ 0.70 & \deltag{0.47}{23} \\

\quad $\mu$Code~\citep{multiturncodegen}   
& 87.81 $\pm$ 2.65 & \underline{94.10 $\pm$ 1.54} & \deltag{0.61}{24}
& 50.87 $\pm$ 0.46 & 71.93 $\pm$ 0.12 & \deltag{1.00}{26} \\

\oursrow
\quad \method\ - \textsc{MaRS}
& \textbf{92.48 $\pm$ 0.61} & \textbf{95.73 $\pm$ 0.31} & \deltag{2.24}{33}
& \textbf{53.33 $\pm$ 1.15} & \textbf{73.33 $\pm$ 1.31} & \deltag{2.40}{34} \\
\bottomrule

\end{tabular}
}
\label{tab:humaneval_mbpp}
\end{table*}

\subsubsection{LiveCodeBench-v6 Per-Difficulty Results}
\label{app:lcb-v6}

Table~\ref{tab:lcb-v6-appendix} reports per-difficulty pass@1 on
LiveCodeBench-v6 using Qwen3-4B as the base model. Base, GRPO-MT,
$\mu$\textsc{Code}, and \textsc{Murphy} are evaluated under the
multi-turn Reflexion framework at 1, 3, and 5 iterations, while
\textsc{ReVeal} is evaluated using its official generation-verification
scaffold with the same inference-turn budgets. This gives \textsc{ReVeal}
its intended evaluation setting: although its generation prompt, output
format, and parser differ from Reflexion, the scaffold is structurally
similar in that it conditions on prior incorrect responses and environment
feedback over multiple turns to refine subsequent attempts. \textsc{Murphy}
demonstrates superior performance in multi-turn (Iter-3/5) settings. On
the Hard subset, gains over the Base model increase with iteration
(+3.57, +5.10, +5.14 pp at Iter-1/3/5). The only exception is the Medium
subset at Iter-1, where $\mu$\textsc{Code} outperforms \textsc{Murphy} by
1.67 pp; however, \textsc{Murphy} surpasses all baselines by Iter-3
(+3.85 pp) and further extends the margin at Iter-5 (+4.38 pp). This
trend supports our claim that \textsc{Murphy}'s training objective
compounds across self-correction iterations.

\begin{table*}[h]
\centering
\caption{\textbf{Per-difficulty pass@1 on LiveCodeBench-v6 with
Qwen3-4B.} Values are mean$_{\pm \text{std}}$ over 3 runs, in
percent. \textbf{Bold} marks the best method per row; \underline{underline}
marks the second best. \textsc{Murphy} achieves competitive or superior performance in multi-turn setting.}

\label{tab:lcb-v6-appendix}
\small
\setlength{\tabcolsep}{5pt}
\renewcommand{\arraystretch}{1.25}
\resizebox{\textwidth}{!}{%
\begin{tabular}{l c ccccc}
\toprule
\textbf{Subset} & \textbf{Iter} & \textbf{Base} & \textbf{GRPO-MT} & \textbf{$\mu$\textsc{Code}} & \textbf{\textsc{ReVeal}} & \textbf{\textsc{Murphy} (Ours)} \\
\midrule
\multirow{3}{*}{Easy}
  & 1 & $84.47_{\pm 2.04}$ & $84.47_{\pm 0.93}$ & $85.09_{\pm 0.62}$ & $\underline{86.23_{\pm 1.09}}$ & $\mathbf{87.49_{\pm 0.52}}$ \\
  & 3 & $92.34_{\pm 1.26}$ & $92.13_{\pm 0.79}$ & $\underline{92.55_{\pm 0.31}}$ & $92.49_{\pm 0.56}$ & $\mathbf{93.38_{\pm 0.18}}$ \\
  & 5 & $93.27_{\pm 0.89}$ & $91.92_{\pm 0.83}$ & $93.62_{\pm 0.18}$ & $\underline{93.79_{\pm 0.54}}$ & $\mathbf{94.23_{\pm 0.16}}$ \\
\midrule
\multirow{3}{*}{Medium}
  & 1 & $45.29_{\pm 3.46}$ & $47.91_{\pm 0.53}$ & $\mathbf{52.79_{\pm 0.66}}$ & $46.51_{\pm 1.53}$ & $\underline{51.12_{\pm 0.53}}$ \\
  & 3 & $58.38_{\pm 1.84}$ & $59.51_{\pm 1.06}$ & $61.23_{\pm 1.28}$ & $\underline{62.38_{\pm 2.58}}$ & $\mathbf{66.23_{\pm 0.54}}$ \\
  & 5 & $61.69_{\pm 1.44}$ & $60.56_{\pm 1.86}$ & $\underline{63.19_{\pm 0.84}}$ & $61.61_{\pm 1.74}$ & $\mathbf{67.57_{\pm 0.61}}$ \\
\midrule
\multirow{3}{*}{Hard}
  & 1 & $10.76_{\pm 0.59}$ & $11.05_{\pm 1.00}$ & $\underline{11.95_{\pm 2.64}}$ & $11.43_{\pm 1.59}$ & $\mathbf{14.33_{\pm 0.41}}$ \\
  & 3 & $16.19_{\pm 1.72}$ & $\underline{18.29_{\pm 0.58}}$ & $18.10_{\pm 1.08}$ & $18.13_{\pm 1.28}$ & $\mathbf{21.29_{\pm 0.43}}$ \\
  & 5 & $19.15_{\pm 1.03}$ & $19.14_{\pm 0.99}$ & $\underline{20.09_{\pm 1.46}}$ & $19.14_{\pm 0.86}$ & $\mathbf{24.29_{\pm 0.38}}$ \\
\bottomrule
\end{tabular}
}
\end{table*}

To test whether the LiveCodeBench-v6 gains observed on Qwen3-4B also hold
for a smaller model, we evaluate Qwen3-1.7B under the same iterative
execution-feedback protocol. As shown in Tab.~\ref{tab:lcbv6_qwen17b},
Iter-1 serves primarily as a sanity check: \method~does not degrade
one-shot generation and achieves the best aggregate performance among all
methods. The main comparison is Iter-3, where models can condition on
execution feedback and refine earlier attempts. In this multi-turn setting,
\method~improves over the strongest non-\method~baseline by +1.93 pp on the
aggregate split and by +3.40 pp on the Medium split. On the Hard split,
MURPHY matches \(\mu\)Code while exhibiting lower variance. These results
show that \method's LiveCodeBench-v6 gains are not specific to Qwen3-4B and
that its advantage is most apparent in the feedback-conditioned multi-turn
setting targeted by our method.

\begin{table*}[h!]
\centering
\caption{
\textbf{LiveCodeBench-v6 results on Qwen3-1.7B.}
We report pass@1 on the Easy, Medium, Hard, and aggregate splits under
1-turn and 3-turn evaluation. Iter-1 serves as a sanity check for one-shot
generation, while Iter-3 is the primary multi-turn setting for evaluating
feedback-conditioned self-correction. \method~achieves the best aggregate
performance at both settings, with the largest gains appearing at Iter-3.
Results are averaged over three runs; standard deviations are shown after
\(\pm\).
}
\label{tab:lcbv6_qwen17b}
\small
\setlength{\tabcolsep}{5.5pt}
\begin{tabular}{c l c c c c}
\toprule
\textbf{Iter.} & \textbf{Variant} & \textbf{Easy} & \textbf{Medium} & \textbf{Hard} & \textbf{Aggregate (All)} \\
\midrule
\multirow{4}{*}{1}
& BASE        & $71.95 \pm 1.09$ & $25.57 \pm 0.66$ & $3.90 \pm 0.33$ & $32.54 \pm 0.19$ \\
& GRPO-MT     & $71.74 \pm 0.54$ & $30.36 \pm 0.46$ & $4.19 \pm 0.33$ & $34.32 \pm 0.24$ \\
& \(\mu\)Code & $73.50 \pm 2.00$ & $33.07 \pm 0.16$ & $4.95 \pm 0.44$ & $36.08 \pm 0.67$ \\
& MURPHY      & $\mathbf{74.43 \pm 1.60}$ & $\mathbf{34.90 \pm 0.66}$ & $\mathbf{6.09 \pm 0.83}$ & $\mathbf{37.41 \pm 0.52}$ \\
\midrule
\multirow{4}{*}{3}
& BASE        & $80.12 \pm 0.94$ & $33.51 \pm 1.71$ & $5.72 \pm 0.50$ & $38.52 \pm 1.06$ \\
& GRPO-MT     & $80.64 \pm 1.77$ & $39.35 \pm 1.23$ & $6.48 \pm 0.59$ & $41.05 \pm 0.05$ \\
& \(\mu\)Code & $82.61 \pm 3.89$ & $42.06 \pm 0.40$ & $\mathbf{8.86 \pm 0.50}$ & $43.42 \pm 1.14$ \\
& MURPHY      & $\mathbf{84.88 \pm 1.00}$ & $\mathbf{45.46 \pm 1.34}$ & $\mathbf{8.86 \pm 0.29}$ & $\mathbf{45.35 \pm 0.81}$ \\
\bottomrule
\end{tabular}
\end{table*}

\subsubsection{Ablation: Effect of Training Dataset Size}
\label{subsec:effect_of_training_data_size}

\begin{figure}[h!]
\centering
\includegraphics[width=\textwidth]{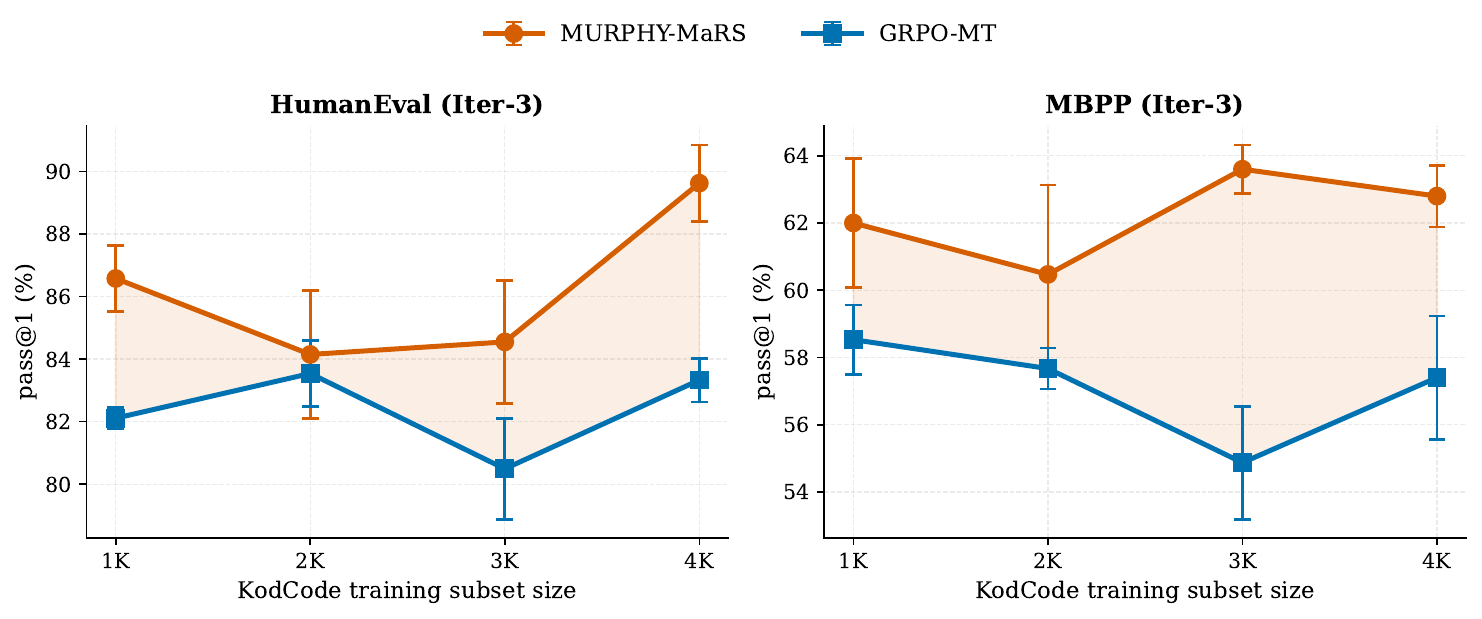}
\caption{\textbf{\method's multi-turn advantage holds across training data 
scales.} Qwen3-1.7B trained on nested KodCode subsets ($1\text{K} \subset 
2\text{K} \subset 3\text{K} \subset 4\text{K}$) and evaluated under 
three-iteration Reflexion (Iter-3) on HumanEval (left) and MBPP (right). 
Markers show the mean over 3 seeds; error bars indicate one standard 
deviation. Shaded regions highlight where \method-\textsc{MaRS} exceeds 
GRPO-MT. \method~outperforms compute-matched GRPO-MT at every dataset 
scale on both benchmarks. We focus on the multi-turn (Iter-3) setting 
here since it directly probes the credit-assignment mechanism that 
distinguishes \method~from GRPO-MT.}
\label{fig:scaling}
\end{figure}

To examine whether \method's gains hold beyond the 1K training split 
used in our main experiments (\autoref{fig:humaneval_mbpp}), we 
construct three additional nested subsets of KodCode ($2\text{K} 
\subset 3\text{K} \subset 4\text{K}$) and train Qwen3-1.7B with both 
GRPO-MT and \method-\textsc{MaRS} under matched compute budgets. 
\autoref{fig:scaling} reports Iter-3 (multi-turn) pass@1 on HumanEval 
and MBPP across all four scales, including the 1K result from 
\autoref{fig:humaneval_mbpp} for completeness. Although neither method 
scales monotonically with dataset size, \method-\textsc{MaRS} 
outperforms GRPO-MT at every scale on both benchmarks, with average 
gains of $+3.9$ pp on HumanEval and $+5.1$ pp on MBPP. This pattern 
indicates that \method's multi-turn advantage is a property of the 
credit-assignment mechanism rather than an artifact of the specific 
training subset used in our main results.

\subsubsection{Ablation: Effect of Pruning in Multi-Turn \method~Training}
\label{app:3_stage}
We noted in the main paper that increasing the number of turns 
in~\method~can lead to exponential growth in computational cost. To 
mitigate this, we design and evaluate two pruning strategies. In this 
experiment, we extend our setup to a 3-turn setting and study the 
effect of pruning on performance. Results in~\autoref{tab:3stage_murphy_pruning} 
show that the pruned variant achieves competitive or even superior 
performance compared to the non-pruned counterpart.
\begin{table*}[h!]
\caption{Ablation study comparing pruning versus non-pruning strategies for Qwen3-1.7B trained with~\method~for 3 turns on the KodCode dataset. Reported numbers indicate \textit{pass@1} (\%) over three independent runs. Pruned variant achieves competitive performance compared to non-pruned~\method. Although the MBPP Iter-3 means round to the same values as the two-turn pruning ablation
in \autoref{tab:murphy_pruning}, these results come from separate three-turn training runs; the different standard deviations reflect different seed-level outcomes.}
\centering
\small
\resizebox{\linewidth}{!}{
\begin{tabular}{lcccc}
\toprule
\multirow{2}{*}{\textbf{Model}} &
\multicolumn{2}{c}{\textbf{HumanEval}} &
\multicolumn{2}{c}{\textbf{MBPP}} \\
\cmidrule(lr){2-3} \cmidrule(lr){4-5}
& \textbf{Iter-1} & \textbf{Iter-3} & \textbf{Iter-1} & \textbf{Iter-3} \\
\midrule
\method - \textsc{MaRS}, \textsc{InterP} & 
$78.66 \pm 2.66$ & $\mathbf{84.15 \pm 1.83}$ & 
$\mathbf{45.53 \pm 1.17}$ & $61.07 \pm 0.42$ \\
\method - \textsc{MaRS} &
$\mathbf{79.27 \pm 2.11}$ & $83.94 \pm 1.41$ & 
$44.80 \pm 1.11$ & $\mathbf{62.20 \pm 1.40}$ \\
\bottomrule
\end{tabular}
}
\label{tab:3stage_murphy_pruning}
\end{table*}

\input{section/prompts}

\section{Efficiency Gains}
\label{app:efficiency_gains}

\paragraph{Practical cost of rollout expansion.}
Although the number of nodes in an unpruned feedback-conditioned tree can
grow with the turn budget, several factors reduce the realized cost in
practice. Successful candidates terminate immediately, and only failed
candidates are expanded. In addition, descendants of the same parent share
substantial prompt prefixes, including the original problem, previous output,
and executor feedback. In our experiments, we use vLLM for rollout
generation~\citep{vllm}. Beyond this setup, the shared-prefix structure could
be further exploited by additional systems techniques such as prefix caching,
asynchronous rollout generation~\citep{primerl}, and speculative or parallel
decoding~\citep{medusa}. These optimizations are not used in our experiments
and are complementary to \method's pruning, which targets post-rollout
optimization cost rather than rollout construction.

\begin{table*}[h!]
\centering
\caption{\textbf{Where pruning saves compute.} Average per-step timing 
breakdown across an end-to-end training run for unpruned \method~vs.\ 
\method~with \textsc{InterP} pruning on Qwen3-1.7B (8$\times$H100s, up to 72 
rollouts per prompt; 8 in turn-1 and up to 64 in turn-2). Pruning is applied 
post-rollout, so generation and reward times are essentially unchanged; the 
speedup concentrates in the optimization phase, where the number of sequences 
entering gradient computation is reduced.}
\label{tab:pruning_timing}
\small
\begin{tabular}{lccc}
\toprule
\textbf{Phase} & \textbf{Unpruned} & \textbf{\textsc{InterP}} & \textbf{$\Delta$} \\
\midrule
Generation (vLLM) & 30.54s & 30.46s & $-0.3\%$ \\
Reward (code exec) & 4.86s & 4.47s & $-8.0\%$ \\
Rollout overhead\textsuperscript{\dag} & 19.86s & 17.75s & $-10.6\%$ \\
\midrule
Rollout total & 55.25s (76\%) & 52.67s (92\%) & $-4.7\%$ \\
Optimization & 17.49s (24\%) & 4.53s (8\%) & $\mathbf{-74.1\%}$ \\
\midrule
\textbf{Total} & \textbf{72.74s} & \textbf{57.20s} & $\mathbf{-21.4\%}$ \\
\bottomrule
\end{tabular}

\vspace{0.3em}
\footnotesize{\textsuperscript{\dag} Reference log-probs, tokenization, and advantage computation.}
\end{table*}

\paragraph{Measured pruning speedup.}
To localize the speedup from pruning, we profile per-step timing averaged
across an end-to-end \textsc{InterP} run against unpruned \method under
matched configurations (Qwen3-1.7B, 8$\times$H100 GPUs, 72 rollouts per
prompt: 8 in turn 1 and up to 64 in turn 2); see~\autoref{tab:pruning_timing}.
Generation with vLLM and reward computation via code execution are
essentially unchanged ($-0.3\%$ and $-8.0\%$, respectively), since pruning is
applied \emph{after} rollouts are collected. The savings instead concentrate
in optimization, where pruning reduces the number of sequences entering
gradient computation: optimization time drops by $\mathbf{74.1\%}$, yielding a
$\mathbf{21.4\%}$ reduction in average per-step training time. This confirms
that \textsc{InterP}'s computational benefit comes from avoiding gradient
updates on uninformative trajectories, not from shortening rollouts. This is
important because aggressive rollout-time pruning could discard feedback
signals that later turns rely on.

\section{Broader Impact}
\label{app:potential_risks}

While \method~introduces some new dynamics through iterative self-correction and reflective optimization, the associated risks appear modest overall. The main considerations involve ensuring that feedback loops remain interpretable and that reward signals do not inadvertently reinforce narrow or heuristic reasoning. There is also some potential for subtle reward hacking, where the model optimizes for easily verifiable but shallow improvements, or for mild distributional drift if reflective heuristics fail to generalize beyond training contexts. Nonetheless, because \method~still relies on verifiable rewards and bounded reflection, these risks are relatively contained and can be mitigated through careful evaluation design, human oversight, and robust validation across diverse task settings.

%% file: section/prompts.tex
\newpage
\section{Prompt Examples}
\label{app:prompt_examples}

To train the \method~objective, we employ the following prompts. The system prompt is used at each dialogue turn, and the feedback prompts are applied during every feedback turn.

\begin{tcolorbox}[
  enhanced,
  breakable,
  colback=ngreen,
  colframe=ngreen,
  coltitle=black,
  title=\textbf{System Prompt},
]
You are a Python coding AI agent. When given a Python function signature and docstring, you must provide a complete Python solution following this exact format:

\begin{enumerate}
  \item \textbf{Reasoning Phase:} Use \texttt{<think>...</think>} tags to contain your complete thought process.
  \begin{itemize}
    \item Break down the problem requirements step by step
    \item Identify key constraints, edge cases, and potential pitfalls
    \item Plan your algorithm and data structures
    \item Walk through examples to validate your approach
    \item Explain your logic thoroughly as if working on scratch paper
  \end{itemize}

  \item \textbf{Implementation Phase:} Use \texttt{<output>...</output>} tags to contain your final code.
  \begin{itemize}
    \item Include the complete function with the original signature
    \item Ensure your code directly implements the approach from your thinking
    \item Write clean, readable code with appropriate comments if needed
  \end{itemize}
\end{enumerate}
Your solution must be complete, correct, and handle all specified requirements.

\end{tcolorbox}

\begin{tcolorbox}[
  enhanced,
  breakable,
  colback=nblue,
  colframe=nblue,
  coltitle=black,
  title=\textbf{Feedback Header},
]
  You have previously attempted this problem \texttt{\{num-attempts\}} time(s).

\end{tcolorbox}

\begin{tcolorbox}[
  enhanced,
  breakable,
  colback=nblue,
  colframe=nblue,
  coltitle=black,
  title=\textbf{Feedback Body},
]

\textbf{Previous Attempt \#\texttt{\{num-attempts\}} Analysis:} \\

Your earlier thought process and implementation: \\

Let me think step by step. \\
\texttt{<think>} \\
\texttt{\{code-generated\}} \\

\textbf{Test Case Results:} \\
- Passed: \texttt{\{passed-tests\}}/\texttt{\{total-tests\}} test cases \\
- Failed: \texttt{\{failed-tests\}}/\texttt{\{total-tests\}} test cases \\

\textbf{Detailed Feedback:} \texttt{\{feedback-string\}} \\

Note: For failed test cases, your code's output is shown followed by a `\#` symbol indicating the failure.

\end{tcolorbox}

\begin{tcolorbox}[
  enhanced,
  breakable,
  colback=nblue,
  colframe=nblue,
  coltitle=black,
  title=\textbf{Feedback Footer.},
]

\textbf{Instructions for Your Next Attempt:}

\begin{enumerate}
  \item \textbf{Failure Analysis Phase:}
    \begin{enumerate}
      \item Carefully examine each failed test case to understand exactly what went wrong
      \item Identify the specific lines of code or logic that caused the failures
      \item Look for patterns across multiple failed cases (e.g., all involve negative numbers, empty inputs, etc.)
      \item Determine if failures stem from algorithmic errors, edge case handling, or implementation bugs
    \end{enumerate}

  \item \textbf{Solution Refinement Phase:}
    \begin{enumerate}
      \item Build upon any correct aspects of your previous solution that passed tests
      \item Redesign the problematic parts of your algorithm to handle the failed cases
      \item Ensure your new approach covers edge cases that weren't properly handled before
      \item Consider additional edge cases that might not be in the test suite but could break your solution
    \end{enumerate}

  \item \textbf{Implementation Phase:}
    \begin{enumerate}
      \item Write your improved solution that directly addresses the identified failure points
      \item Test your logic mentally against the failed cases to verify it would now pass
      \item Ensure your solution maintains correctness for previously passing test cases
    \end{enumerate}
\end{enumerate}

Use the concrete feedback from your \texttt{\{num-attempts\}} previous attempt(s) to create a more robust and accurate solution.

\end{tcolorbox}